\documentclass{article}

\usepackage[preprint]{neurips_2026}

\usepackage[utf8]{inputenc}
\usepackage[T1]{fontenc}
\usepackage{hyperref}
\hypersetup{hidelinks}
\usepackage{url}
\usepackage{booktabs}
\usepackage{amsfonts}
\usepackage{amsmath}
\usepackage{nicefrac}
\usepackage{microtype}
\usepackage{xcolor}
\usepackage{graphicx}

\usepackage{tikz}
\usetikzlibrary{positioning, shapes.geometric, arrows.meta, fit, backgrounds, calc}

\usepackage{enumitem}
\usepackage{comment}
\usepackage{placeins}

\newcommand{\benchname}{AGC-Bench}
\newcommand{\benchmarkcount}{78}
\newcommand{\textonlycount}{67}
\newcommand{\multimodalcount}{11}

\newcommand{\finalistcount}{83}
\newcommand{\domaincount}{six}
\newcommand{\domaincountnr}{6}
\newcommand{\projecturl}{\url{https://huggingface.co/agcbench-2026}}

\title{\benchname: Measuring Artificial General Creativity}

\author{%
  \textbf{Roger Beaty}$^{*, 1}$ \quad
  \textbf{Vijeta Deshpande}$^{*, 2}$ \quad
  \textbf{Clin K.Y. Lai}$^{1}$ \quad
  \textbf{Anna Attuch}$^{3}$ \quad
  \textbf{Namrata Shivagunde}$^{2}$ \\
  \textbf{Swastik Roy}$^{4}$ \quad
  \textbf{Rajkumar Pujari}$^{4}$ \quad
  \textbf{Paul V. DiStefano}$^{1}$ \quad
  \textbf{Sherin Muckatira}$^{2}$ \\
  \textbf{Claire E. Stevenson}$^{3}$ \quad
  \textbf{Mikhail Gronas}$^{5}$ \quad
  \textbf{Anna Rumshisky}$^{2,4}$ \\[0.5em]
  $^{1}$Pennsylvania State University \quad
  $^{2}$University of Massachusetts Lowell \\
  $^{3}$University of Amsterdam \quad
  $^{4}$Amazon AGI \quad
  $^{5}$Dartmouth College \\[0.3em]
  {\small $^{*}$Equal contribution}
}

\begin{document}

\maketitle

\begin{abstract}
Creativity research has long debated whether creativity is a general
ability or specific to particular domains (e.g., creative writing,
visual art, scientific discovery), and whether it is separable from
general intelligence. Both questions now apply to LLMs, but a
fragmented evaluation landscape across hundreds of heterogeneous
creativity benchmarks has left them empirically intractable. We
introduce \textbf{\benchname}, a meta-benchmark for artificial
general creativity built from a PRISMA-compliant systematic review
of the AI creativity literature ($3{,}101$ papers screened, $497$
unique benchmarks identified) paired with an agentic onboarding
harness that converts source-paper benchmarks into runnable
HELM-style scenarios. The first release covers \benchmarkcount{}
datasets (\textonlycount{} text-only, \multimodalcount{} multimodal)
spanning brainstorming, problem solving, STEM, narrative, figurative
language, and humor. To address bias in LLM-as-judge, we
apply Judge Response Theory (JRT)---a psychometric calibration of
judge leniency/severity---with three frontier LLM judges. We then
fine-tune \texttt{Qwen3-30B-A3B-Instruct-2507} on the resulting
$48{,}299$ JRT-corrected ratings to produce \textbf{AGC-Judge}, an
open-weight scoring model that matches the three-judge ensemble and
predicts frontier-judge ratings with high accuracy on creativity
benchmarks it was not trained on. Results reveal frontier models at
the top of the leaderboard, with open-weight models close behind.
However, different models show different creative strengths, ranking
higher on some domains (e.g., creative writing) than others (e.g.,
scientific ideation). We conduct several experiments and report
three main findings. First, applying factor analysis across $83$
frontier LLMs, we recover a single creativity factor `\textit{c}',
analogous to the `\textit{g}' factor of general intelligence, that
explains $81.5\%$ of variance, related to but separable from general
knowledge and reasoning. Second, in within-model
comparisons, we show that prompting models to ``be creative'' boosts
their performance far more than enabling their reasoning capability,
evidence that the benchmark tracks creativity specifically. Third,
on a human-matched subset of five tasks, we find creativity is
significantly more domain-general in LLMs than in humans, although
the top human still leads the top LLM. We release
\benchname{} with a public leaderboard, the onboarding harness,
AGC-Judge weights, and human comparison data as open infrastructure
for measuring AI creativity at scale.
\end{abstract}

\begin{figure}[!t]
\centering
\includegraphics[width=\linewidth]{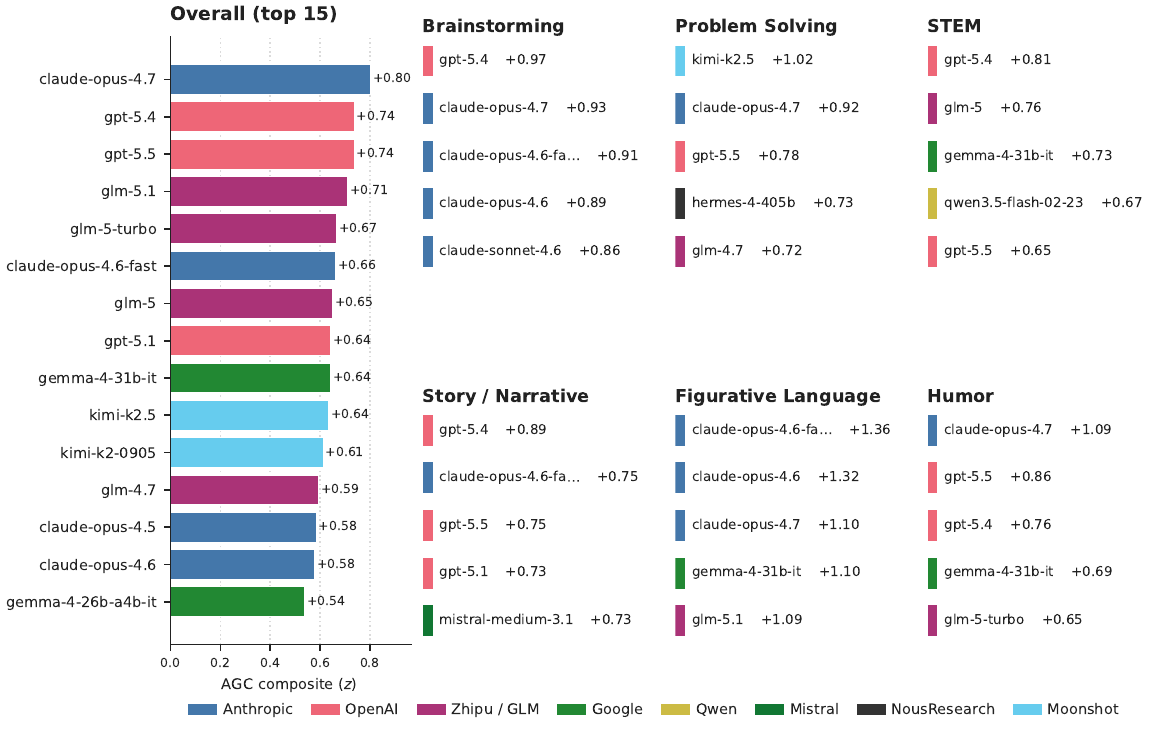}
\caption{\textbf{Overall and per-domain rankings.} \textbf{Left:}
top 15 of \finalistcount{} models on the \benchname{} composite.
\textbf{Right:} top 5 within each of the six domains. Per-domain
lineups overlap substantially with the overall ranking but reorder at
the margins (e.g., \texttt{kimi-k2.5} on Problem Solving,
\texttt{claude-opus-4.6-fast} on Figurative Language). Full
\finalistcount{}-model leaderboard in Appendix~\ref{app:models}.}
\label{fig:leaderboard}
\end{figure}

\section{Introduction}

Creativity is widely regarded as a hallmark of human intelligence, and
large language models (LLMs) are increasingly being asked to produce
creative output---writing stories, generating research ideas, designing
marketing copy, composing poetry \citep{peeperkorn2025divergent,
bhat2025creativity, ruan2024liveideabench}. Evaluating how well they do
it, however, remains challenging. Hundreds of creativity benchmarks
have been published in recent years, each targeting a narrow slice of
creative behavior, each using its own scoring protocol, and most run
in isolation. Yet the field has no infrastructure for asking whether a
model's creative ability generalizes across domains---whether a model
that writes compelling short stories also generates clever metaphors,
designs novel product taglines, or formulates original scientific
hypotheses.

The same question is non-trivial in humans. Previous work suggests that creative performance
reflects both general abilities that support many kinds of creative
work and domain-specific abilities tied to particular tasks
\citep{baer2010neglect, plucker2004generality, kaufman2009beyond}.
Someone who writes good poetry, for example, does not need to be equally inventive when
sketching or solving engineering problems. Whether a similar pattern
holds for LLMs is unknown, in part because the field lacks the
evaluation infrastructure to answer it. We call this open question, by analogy with artificial general
intelligence, the problem of \textit{artificial general creativity}:
does creative performance in LLMs reflect a single general capacity
that transfers across domains, or a set of domain-specific abilities
shaped by training data, architecture, and post-training objectives?
To address this question, we take a psychometric approach. Creativity
researchers have long used psychometric tools, such as factor analysis
and item response theory, to test whether human creative ability is
general or domain-specific. We apply the same tools to large-scale
evaluation data from LLMs. To do so, we need a unified framework that
measures creative ability across diverse domains.

We introduce \benchname{}, a meta-benchmark built by aggregating the
published literature on AI creativity evaluation. A PRISMA-compliant
systematic review \citep{page2021prisma} of \textbf{3{,}101 candidate
papers} yielded \textbf{497 unique benchmarks} published between 2018
and 2025. We onboarded \benchmarkcount{} of them into a standardized
evaluation harness modeled on HELM \citep{liang2023helm}, organized
around \domaincount{} text-only domains drawn from human creativity
research---brainstorming, problem solving, STEM, story / narrative
creativity, figurative language, and humor---as well as a multimodal
visual and design subset. We evaluated \finalistcount{} models spanning diverse model families
and scales. Each benchmark was scored using its original metric. To
address bias in LLM-as-judge scoring, we applied Judge Response
Theory (JRT) \citep{myszkowski2019judge}---a psychometric calibration
of each judge's leniency and severity in rating creativity---with a three-frontier-vendor panel
(Gemini-3-Flash, Grok-4.1-Fast, GPT-4.1-mini) under a planned-missing
2-of-3 design. We used the resulting calibrated estimates to train
AGC-Judge, an open-weight creativity rater described below. To test
whether creative ability in LLMs is general or domain-specific, we
aggregated per-model scores across the \textonlycount{} text-only
datasets into \domaincount{} domain composites and ran factor
analysis to extract a general creativity factor (\textit{C}).

We report three main findings. \textbf{First}, across
\finalistcount{} frontier LLMs, factor analysis reveals a general
\textit{C} factor explaining 81.5\% of the variance across
the \domaincount{} text-only domains ($\alpha = 0.96$); the factor
correlates with general knowledge and reasoning yet remains separable
from each. \textbf{Second}, in within-model comparisons, prompting
models to ``be creative'' boosts their creative performance far
more than enabling reasoning, evidence that the benchmark tracks
creativity rather than reasoning. \textbf{Third}, on a five-task
subset with paired human responses, creativity is more domain-general
in LLMs than in humans, although the top human still leads
the top LLM, with frontier models approaching but not yet matching
the strongest human performers.

This work contributes three primary artifacts:
\begin{itemize}[noitemsep, topsep=0pt, partopsep=0pt]
\item \textbf{AGC-Bench.} A benchmark for artificial general creativity
built from \benchmarkcount{} onboarded datasets (\textonlycount{}
text-only, \multimodalcount{} multimodal). Each dataset runs as a
HELM-compatible scenario scored with the canonical metric from its
source paper. We release the harness, model-output generations, and a
public leaderboard.
\item \textbf{AGC-Judge.} An open-weight Qwen3-30B-A3B scoring model
fine-tuned on \textbf{48{,}299} calibrated three-judge ratings.
AGC-Judge matches the three-judge ensemble at Spearman
$\rho \geq 0.97$ in-distribution and predicts frontier-judge ratings
on creativity benchmarks it was not trained on at $\rho = 0.83$, at a
fraction of frontier-judge cost.
\item \textbf{Meta-benchmarking pipeline.} A reproducible workflow
that converts source-paper benchmarks into runnable scenarios,
combining a PRISMA-compliant systematic review with an agentic
onboarding harness. The pipeline supports incremental coverage of the
literature beyond this release, and extends naturally to
building meta-benchmarks in other domains.
\end{itemize}
We also release the full PRISMA-curated inventory of \textbf{497
creativity benchmarks} surveyed during onboarding---a superset of the
\benchmarkcount{} we run, annotated with modality, scoring protocol,
and source paper, supporting future expansion of the harness.
\benchname{} constitutes an order-of-magnitude larger creativity
benchmark than prior work and an initial empirical test of the
artificial general creativity question. All artifacts are available at
\projecturl.

\section{Related Work}
\label{sec:related}

\paragraph{General and domain-specific creativity.}
A central question in human creativity research is whether creative
ability reflects a single general capacity, analogous to Spearman's
$g$ for intelligence, or whether creativity is domain-specific.
Large-scale assessments of creative behavior, including the Creative
Achievement Questionnaire \citep{carson2005reliability}, the Kaufman
Domains of Creativity Scale \citep{kaufman2012beyond}, and the
Inventory of Creative Activities and Achievements
\citep{diedrich2018assessment}, consistently partition creative
performance across multiple content domains---visual arts, creative
writing, humor, scientific and technical creativity, and everyday
problem-solving---with moderate correlations across domains and
substantial domain-specific variance \citep{baer2010neglect,
kaufman2009beyond}. Creative performance is typically
assessed using tasks that require the generation of ideas or products that are both
novel and appropriate \citep{runco2012standard, amabile1996creativity}.
\benchname{} adopts this framework directly. The \domaincount{}
domains map onto content categories that recur across these
taxonomies, paired with process-level traditions (divergent thinking
and creative problem solving) that creativity researchers have long studied
alongside them. Model performance can therefore be situated within
the same structure used to characterize human creative ability.

\paragraph{Intelligence, creativity, and implications for LLMs.}
The relationship between intelligence and creativity has a long and
contested history in psychology, with direct implications for
the evaluation of LLMs. Within the Cattell-Horn-Carroll framework
\citep{mcgrew2009chc}, intelligence comprises a general factor ($g$)
and broad abilities including fluid intelligence ($G_f$, the capacity
for novel problem solving through inductive and deductive reasoning)
and crystallized intelligence ($G_c$, accumulated declarative and
procedural knowledge). $G_f$ supports the generation and evaluation
of candidate ideas, $G_c$ provides the knowledge base from which they
are constructed, and both contribute to creative thinking
 \citep{gerwig2021intelligence}. Frontier LLMs
exhibit performance analogous to both components, and many
existing ``creativity'' benchmarks plausibly index $G_f$ in
disguise. Distinguishing creative ability from reasoning therefore
requires a targeted $G_f$ indicator that resists mere pattern-matching. \citet{lewis2024evaluating} construct counterfactual
letter-string analogies, replacing the Latin
alphabet with shuffled alphabets and symbol sequences to force rule
abstraction over memorization---a logic paralleled by Chollet's
Abstraction and Reasoning Corpus \citep{chollet2019measure}. We adopt
the counterfactual letter-string task as our pure-$G_f$ probe
alongside the \benchname{} composite.

\paragraph{The LLM creativity benchmark landscape.}
Benchmarks evaluating creative capability in LLMs have proliferated
rapidly across multiple lines of work, with each subfield developing
its own conventions in relative isolation. The earliest and largest
group adapts the classical psychometric tasks of divergent
thinking---the Alternate Uses Task, the Torrance Tests, the Divergent
Association Task---to LLMs, generally reporting that frontier models
match or exceed average human performance while falling short of the
most creative humans \citep{stevenson2022putting, guzik2023chatgpt,
hubert2024current, bellemarepepin2026divergent}. Creative writing has
 become the most active area, with benchmarks that differ in length,
granularity, and evaluation philosophy and range from constrained
short-story comparisons \citep{gomez2023confederacy} to expert-rubric
evaluation \citep{chakrabarty2024art}. Humor evaluation has
consolidated around image captioning such as the New Yorker Caption
Contest \citep{hessel2023androids}, with extensions targeting humor
reasoning \citep{horvitz2025humorbench} and expert-aligned ranking
\citep{zhou2025bridging}. Other lines of work address figurative
language and metaphor \citep{chakrabarty2023spy, liu2022figqa},
scientific ideation \citep{ruan2024liveideabench}, visual creativity
\citep{chen2025creation}, human-versus-AI creative drawing
\citep{nath2025pencils}, and ``lateral thinking'' through
puzzle-solving tasks \citep{jiang2023brainteaser}. Each benchmark
deepens understanding of one creative domain, but because they were
built independently---different prompts, scoring protocols, and model
rosters---their results cannot be combined to ask broader questions
about a model's creative profile across domains.

\paragraph{Multi-domain evaluation and the present work.}
A few recent efforts have begun to bridge across domains. The closest
prior consolidation is CreativityPrism \citep{hou2025creativityprism},
which assembles nine tasks across divergent thinking, creative
writing, and logical reasoning and evaluates seventeen
state-of-the-art models with a single LLM judge. The authors report
that performance is more correlated within than across their
three domains, suggesting that creative ability in LLMs may not
generalize. \benchname{} extends this line on three dimensions: scale
(\textonlycount{} text-only benchmarks rather than nine and
\finalistcount{} frontier models rather than seventeen, with coverage
of the major 2025--2026 frontier releases that postdate the
CreativityPrism release), scoring methodology (a 3-judge
planned-missing design with Bayesian rater calibration rather than a
single LLM judge), and psychometric grounding (formal factor analysis
with parallel analysis on an $\finalistcount{} \times \domaincountnr{}$
matrix rather than per-pair correlations on the $17 \times 9$ matrix
their cohort supports). Six of the nine CreativityPrism tasks
(\texttt{ttcw}, \texttt{dat}, \texttt{cs4}, \texttt{neocoder},
\texttt{creativemath}, and a 3-word-cue short-story analog) appear
among the \textonlycount{} benchmarks we evaluated. Other work has
examined the validity of the field's evaluation practices and metrics.
\citet{zhao2025rethinking} catalog systematic confounds in existing
creativity benchmarks, e.g., prompt sensitivity, model inconsistency,
and potential data contamination, arguing that even within a single
benchmark, reported scores may not robustly reflect underlying
creative ability. \citet{bhat2025creativity}, the most rigorous
single-domain study to date, found tightly clustered model
performance and poor agreement between LLM judges and human experts
on a large pairwise marketing-creativity comparison. Taken together,
prior work indicates that creative performance is heterogeneous
across domains and tasks, and that characterizing LLM creativity requires
breadth that no single benchmark can provide. \benchname{} addresses
this gap by combining \textonlycount{} primary-cohort benchmarks
across \domaincount{} text-only creative domains, with
\multimodalcount{} additional multimodal scenarios released as
artifacts, and a domain taxonomy drawn from human creativity
research and benchmark-level factor analysis as a formal test of the
\textit{artificial-general-creativity hypothesis}.

\section{Methods}
\label{sec:methods}

\benchname{} contains \textonlycount{} text-only datasets onboarded from our systematic review, with each dataset scored using the metric from its source paper. The present submission is the first release of an ongoing onboarding effort; the harness is designed to extend to additional benchmarks as they are integrated. The release also includes a human subset, \textit{AGC-Human}, of five tasks from the Creativity Assessment Platform \citep{patterson2025cap} that place LLMs and humans on directly comparable scales.

\subsection{Benchmark Curation}
\label{sec:curation}

We assembled the benchmark inventory through a PRISMA-compliant
\citep{page2021prisma} systematic literature review---the gold-standard
evidence-synthesis protocol from medicine, applied here to the AI
creativity literature---with six reproducible stages: literature
harvesting, automated pre-screening, dataset verification, dual human
review, benchmark extraction, and deduplication. The pipeline
interleaves LLM automation for triage at scale with human review at
every accept step. The flow of papers and benchmarks through the
pipeline appears in Figure~\ref{fig:prisma}.

\paragraph{Pre-screening.}
From an initial pool of 3{,}101 candidate papers (January 2018--December 2025), \texttt{gpt-4.1} applied
a structured inclusion rubric to each title and abstract. A paper was
retained if it (a) targeted an AI or LLM system, (b) introduced or
substantially modified a benchmark or dataset, (c) focused on a
creativity-related construct (e.g., ideation, originality, humor,
metaphor, discovery), (d) included either a generation task with
multiple valid solutions or a judgment task rating creative outputs,
and (e) was published between 2018 and 2025. Survey papers without a
new benchmark and papers where creativity was incidental were
excluded. The rubric retained 683 papers.
\texttt{gemini-2.5-flash} with Google Search grounding then
verified that 431 had publicly accessible code or data. Four trained
graduate students then reviewed the resulting candidate set under a
rotating assignment protocol, with conflict resolution by the lead
authors.

\paragraph{Extraction and deduplication.}
Of the 431 candidates with verified data access, 283 passed the dual
human review and were forwarded to extraction. Benchmark extraction
with \texttt{gemini-2.5-pro} parsed the 276 papers with
retrievable PDFs into structured records, producing 546 candidate
benchmarks across 1{,}160 constituent tasks. Each record captured
benchmark name, modality, task description, evaluation metric, and
dataset URL. A hybrid pipeline---string-normalization on
canonicalized names, followed by 10-pass
\texttt{gemini-2.5-pro} voting deduplication with $\geq$80\%
inter-pass agreement---identified 33 duplicate groups, yielding 497
unique benchmarks.

\paragraph{Candidate selection.}
The 497 unique benchmarks underwent a relevance pass that removed
general NLP and computer-vision co-extractions miscategorized by
upstream extraction, leaving 432 creativity-relevant candidates. Each
candidate was then classified by modality and infrastructure
requirements: \texttt{gemini-3-flash-preview} assigned a tier
based on input/output type, with the lead authors reviewing
tier-boundary cases. The classifier yielded 241 Tier-1a (text-to-text,
standard API calls) and 86 Tier-1b (image-to-text, vision APIs),
leaving 105 higher-tier candidates. Tier-1a and selected Tier-1b
candidates entered the onboarding pipeline
(Section~\ref{sec:onboarding}). Higher-tier candidates are retained
in the catalog for future releases.

\subsection{Benchmark Onboarding}
\label{sec:onboarding}

The 327 API-runnable Tier-1a and Tier-1b candidates ship in
heterogeneous formats and use incompatible evaluation paradigms. An
agentic onboarding workflow (\texttt{benchmark-onboarder}) built with
Claude Code on \texttt{claude-opus-4.6} coordinates a
nine-state pipeline from paper parsing through scenario generation
and pilot execution, with three design principles---verbatim prompt
extraction, source-paper scoring metrics preserved, and
regex/code-based extraction preceding LLM inference---and human
review at every accept step. The pipeline produced 159
HELM-compatible scenarios; after a launch-readiness audit,
\benchmarkcount{} entered the present evaluation cycle
(\textonlycount{} text-only and \multimodalcount{} multimodal), with
the remaining scenarios either documented as excluded for scope or
methodology reasons (Appendix~\ref{app:taxonomy}) or queued for
future evaluation. Failure modes and the full state-machine specification are
documented in Appendix~\ref{app:onboarding}.

\subsection{Benchmark Taxonomy}
\label{sec:taxonomy}

The \textonlycount{} text-only datasets are organized into a
taxonomy grounded in human creativity research, with six text-only
domains mapped onto content categories shared across major
self-report inventories of creative behavior
\citep{carson2005reliability,kaufman2012beyond,diedrich2018assessment}:
\textit{Brainstorming}, \textit{Problem Solving}, \textit{STEM},
\textit{Story / Narrative}, \textit{Figurative Language}, and
\textit{Humor}.
Domain assignments come from a three-model LLM panel
(\texttt{gemini-3-flash-preview}, \texttt{grok-4.1-fast},
\texttt{gpt-4.1-mini}) on benchmark metadata, adjudicated by
majority vote with lead-author resolution (mean pairwise Cohen's
$\kappa \approx 0.85$, Fleiss' $\kappa \approx 0.85$ on the released
taxonomy panel); per-dataset assignments, per-domain counts, and the
partition rationale appear in Appendix~\ref{app:taxonomy}.

\paragraph{Scope of the present analysis.}
The first \benchname{} release covers \benchmarkcount{} datasets:
\textonlycount{} primary-cohort text-evaluable datasets (one
image+text figurative-language dataset, IRFL, is included;
its inclusion shifts model composites by $\leq 0.04$ z) and
\multimodalcount{} multimodal scenarios released as artifacts. All
analyses reported in Section~\ref{sec:results} are restricted to the
\textonlycount{} primary-cohort datasets, on which both the
strict-coverage model selection and the JRT calibration were
performed.
\textit{The \multimodalcount{} multimodal scenarios are released as
artifacts for use by the community, but no multimodal results are reported in the
present paper.}



\subsection{Evaluation Setup}
\label{sec:eval}

\paragraph{Harness and sampling.} 
Our evaluation harness extends HELM \citep{liang2023helm} and
lm-evaluation-harness \citep{gao2024harness}. Each benchmark is
implemented as a self-contained scenario with a stable identifier. We
evaluate \finalistcount{} text-only models released between 2024 and
2026, drawn from the major proprietary and open-weight families and
run via vendor APIs and OpenRouter with reasoning off by default. To
make a model $\times$ benchmark grid of this size tractable, each
(model, dataset) cell scores $n = 50$ items sampled with a frozen
seed from the source benchmark (smaller datasets run exhaustively,
canonical splits preserved). This cap prioritizes breadth across
benchmarks and domains over depth on any single one. A rank-stability
bootstrap confirms that leaderboard rankings stabilize well below
this cutoff: Spearman $\rho = 0.99$ between $k = 10$ and the full
$k = 50$, with Top-10 Jaccard $= 1.0$ by $k = 50$ (per-model coverage
and the full bootstrap in Appendix~\ref{app:onboarding}).


\paragraph{Scoring and aggregation.}
Each dataset is scored with its canonical metric, with the
source-paper rubric and judge prompt preserved verbatim
(Appendix~\ref{app:metric_types}). For the 24 LLM-judge benchmarks,
the cohort-scaling judge is \texttt{gemini-3-flash-preview} following
HELM precedent \citep{liang2023helm}, with two cross-vendor judges
(\texttt{grok-4.1-fast}, \texttt{gpt-4.1-mini}) added per cell under
a planned-missing 2-of-3 design. To correct systematic differences in
severity and discrimination across judges, we fit a Bayesian graded
response model from the Judge Response Theory framework
\citep{myszkowski2019judge}, which recovers per-unit ability $\theta$
alongside per-judge discrimination and severity. The corrected scores
preserve the leaderboard rank order while flattening per-judge scale
differences (Figure~\ref{fig:jrt}; full specification and
rank-preservation diagnostics in Appendix~\ref{app:jrt}). The
\benchname{} composite is the per-model mean of dataset-level
z-scores across the \textonlycount{} text-only datasets.
Multi-metric datasets are mean-aggregated, and domain composites are
computed analogously for the $C$-factor analysis
(Section~\ref{sec:cfactor}).

\paragraph{Data-quality audit and reproducibility.}
\label{sec:dq}
An independent LLM-judge audited three random instances per cell.
On-task rate was $95.1\%$, with second-judge overlap of
$\kappa = 0.67$ (Appendix~\ref{app:dq}). The catalog, harness,
responses, judge verdicts, and a public submission leaderboard are
released at \projecturl{}, with each scenario versioned and
traceable to its source paper.

\paragraph{External validity.}
\label{sec:intervention}
Three external designs anchor the construct \benchname{} measures
(full specifications in Appendices~\ref{app:intervention}
and~\ref{app:aa_battery}): the Lewis--Mitchell counterfactual
letter-string analogy task \citep{lewis2024evaluating} as a
pure-$G_f$ discriminant probe; a 1{,}862-item subset of the MuCE
creativity battery \citep{ismayilzada2025crpo} as an out-of-cohort
convergent reference against human ratings; and two within-model
interventions (be-creative vs.\ be-effective prompting; reasoning-mode
toggle) testing whether the composite shifts more under
creativity-targeted than reasoning-targeted manipulation.


\subsection{The AGC-Judge scorer}
\label{sec:jrt}

We fine-tuned \texttt{Qwen3-30B-A3B-Instruct-2507} as a LoRA adapter
to produce \textbf{AGC-Judge}, an open-weight creativity scorer that
reproduces the JRT-calibrated three-judge ensemble at a fraction of
the cost. The training corpus comprises 48{,}299 (model, item)
ratings drawn from the JRT-corrected pool, with each row pairing a
model response to a JRT-derived integer rating on the originating
benchmark's scale, and the adapter is trained with the standard
causal-LM objective. We release the weights and the scoring harness
so others can score new models against the \benchname{} leaderboard
without running a frontier-judge ensemble (full training
configuration, hyperparameters, length-filtering, and ablations in
Appendix~\ref{app:jrt:agcjudge}; held-out evaluation in
Section~\ref{sec:agc_judge_results}).

\subsection{Paired-human subset (AGC-Human)}
\label{sec:cap_human_ai}

AGC-Human places humans and LLMs on directly comparable scales across
five tasks from the Creativity Assessment Platform
\citep{patterson2025cap}: three idea-generation tasks (Alternate
Uses, Scientific Creative Thinking, Design Problems) and two
creative-writing tasks (Metaphor Completion, Short Story). The cohort comprises 224 humans and 125 LLM evaluations. An LLM-judge
validity gate filters out off-task or non-responsive entries (pass
rates $96.0\%$ for humans, $98.8\%$ for LLMs), leaving 201 humans
and 80 LLMs in the canonical-analysis subset. Each response is then
scored on novelty \citep{johnson2022dsi} and diversity using embedding models, with scores
length-residualized and z-scored within (task, prompt). Task
definitions and validity filtering appear in Appendix~\ref{app:intervention}.

\section{Results}
\label{sec:results}

\subsection{Frontier models lead overall, but no model leads every domain}
\label{sec:rankings}

The left panel of Figure 1 shows the top 15 models
on the overall composite. Proprietary frontier systems lead the
field---\texttt{anthropic/claude-opus-4.7},
\texttt{openai/gpt-5.4}, \texttt{openai/gpt-5.5}---with
open-weight \texttt{z-ai/glm-5.1} and \texttt{glm-5-turbo} close
behind. The right panel of Figure 1 reorders the same
models by per-domain composite and surfaces dissociations that diverge from the
overall ranking. \texttt{kimi-k2.5} tops Problem Solving
(+1.02) without entering any other domain top-5.
\texttt{claude-opus-4.6-fast} leads Figurative Language by a wide
margin (+1.36). Open-weight \texttt{gemma-4-31b-it} enters three
domain top-fives (STEM, Figurative Language, Humor) despite ranking
ninth overall. Creative ability is therefore unevenly distributed
even among top models, a pattern we examine quantitatively in
Section~\ref{sec:cfactor}.

\subsection{AGC-Judge generalizes across held-out splits at frontier-judge accuracy}
\label{sec:agc_judge_results}

\begin{figure}[!t]
\centering
\includegraphics[width=0.92\linewidth]{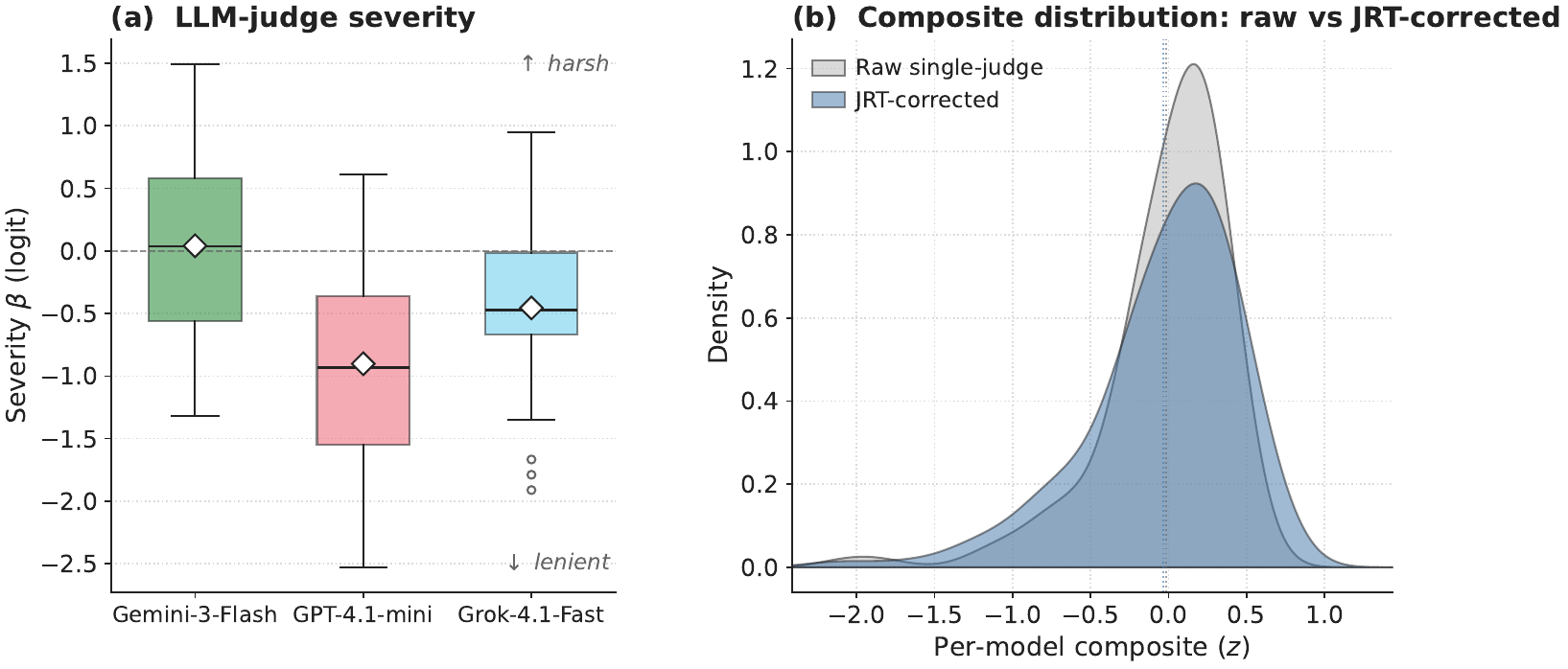}
\caption{\textbf{Judge-response-theory (JRT) calibration of the LLM-judge benchmarks.}
\textbf{(a)} Per-judge severity $\beta$ across 24 LLM-judge cells:
\texttt{gpt-4.1-mini} runs substantially lenient (median
$\beta = -0.93$); \texttt{gemini-3-flash} sits near zero.
\textbf{(b)} Per-model composite density: the raw single-judge
composite (gray) is left-skewed; JRT correction (blue) yields an
approximately normal distribution.}
\label{fig:jrt}
\end{figure}

\begin{figure}[!t]
\centering
\includegraphics[width=\linewidth]{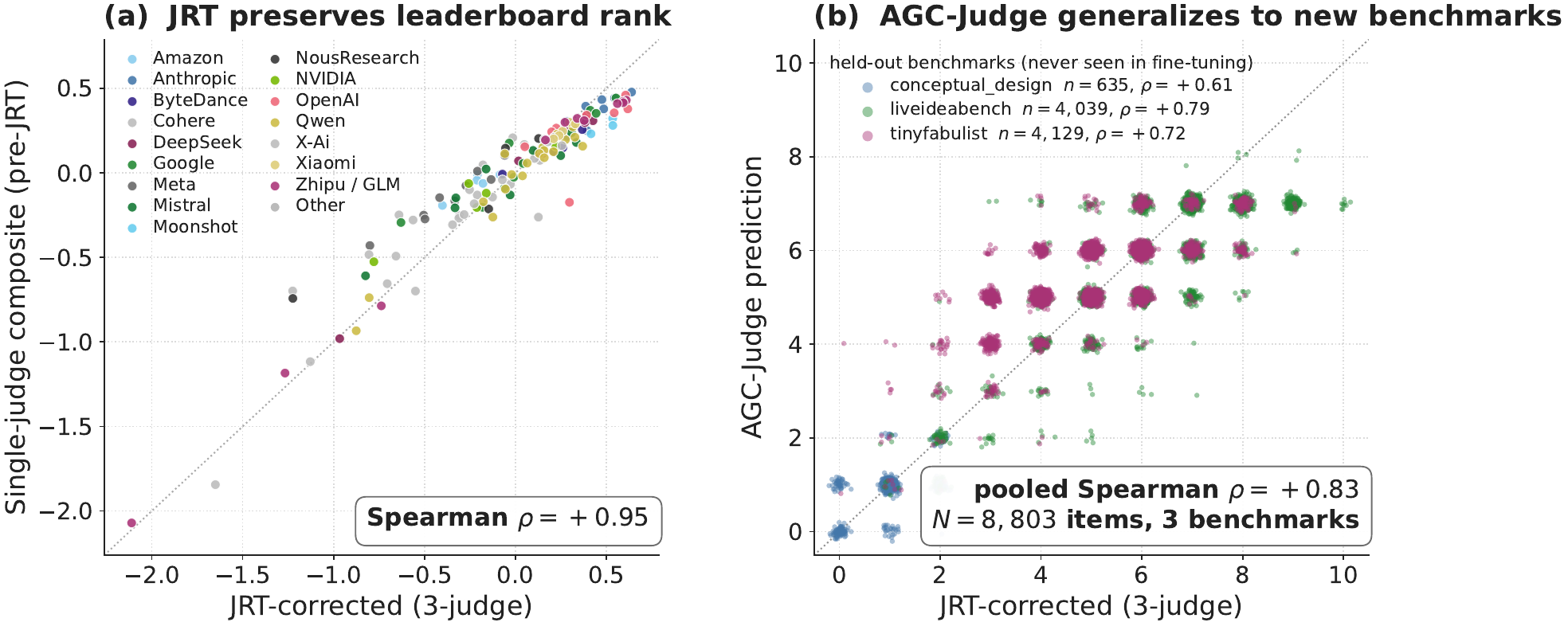}
\caption{\textbf{JRT calibration validation and AGC-Judge generalization.}
\textbf{(a)} JRT-corrected and single-judge canonical composites
cluster on the diagonal at Spearman $\rho = +0.95$ across
\finalistcount{} models---calibration preserves leaderboard rank.
\textbf{(b)} AGC-Judge (open-weight Qwen3-30B LoRA fine-tuned on
$48{,}299$ JRT-corrected ratings) generalizes to three held-out
creativity benchmarks at pooled Spearman $\rho = +0.83$ on $8{,}803$
items.}
\label{fig:judge_generalization}
\end{figure}

We test whether the calibrated three-judge ensemble can be
approximated by a single fine-tuned open-weight scorer. We find that AGC-Judge---the LoRA fine-tune of \texttt{Qwen3-30B-A3B-Instruct-2507} on the
$48{,}299$ JRT-corrected ratings---predicts JRT-corrected scores at
Spearman $\rho = 0.94$ on held-out items and frontier models and
$\rho = 0.83$ on three held-out benchmarks
(Figure~\ref{fig:judge_generalization}), matching the three-judge
ensemble at $\rho \geq 0.97$. On an
out-of-distribution corpus of creative writing from humans and LLMs \citep{orwig2024language}, AGC-Judge
tracks human ratings comparably to a frontier GPT judge with
substantially lower LLM-preference bias
(Appendix~\ref{app:jrt:orwig}). AGC-Judge thus provides a low-cost, robust, and human-aligned creativity scoring model for \benchname{}.

\subsection{A general creativity factor ($C$) in LLMs}
\label{sec:cfactor}

We next ask whether creative ability in LLMs is general across the
six domains or specific to each. To this end, we conducted a factor
analysis on the $83 \times 6$ per-(model, domain) composite matrix. Results show that the structure is clearly unidimensional. The first
eigenvalue is $4.89$---the only one above Kaiser, with the remaining
five at $0.47$ or below---and parallel analysis on 1{,}000 random
matrices \citep{horn1965rationale} places it well above the noise
floor. The resulting factor, which we call \textit{C} by analogy
with \textit{g} for general intelligence, explains $81.5\%$ of
variance, with all six domain loadings between $+0.87$ and $+0.94$
($\alpha = 0.96$). Thus, knowing how a model performs in one
creative domain reliably predicts its performance in the others.

Critically, the factor survives three independent threats to its
validity. First, it holds after partialling out fluid reasoning
($\alpha = 0.91$ on the letter-string-analogy-residualized matrix), so \textit{C} is
not a re-parameterization of $G_f$. Second, it holds within scoring
family: re-extraction on the formula-based subset alone
($\alpha = 0.91$) and the LLM-judge subset alone ($\alpha = 0.96$)
each yields a unidimensional solution, so \textit{C} is not an
artifact of any single scoring class. Finally, \textit{C} is stable
across model subsets: bootstrap subsampling to 60 of the 83 models
gives a consistent eigenvalue ($4.92$, $95\%$ interval
$[4.74, 5.05]$), and dropping the largest domain (Story/Narrative,
20 datasets) leaves the structure intact ($\alpha = 0.94$). Full
robustness diagnostics appear in Appendix~\ref{app:cfactor_robustness}.

\subsection{\textit{C} correlates with intelligence yet remains separable}
\label{sec:c_intelligence}

A unitary \textit{C} would not be a meaningful psychometric finding
if creativity in LLMs simply reduced to general capability. We
therefore tested three competing reductions: \textit{C} as fluid
reasoning, \textit{C} as crystallized knowledge, and \textit{C} as
raw model scale. 

We find that \textit{C} correlates substantially with both components of human
intelligence, but separably. The \benchname{} composite correlates
with the fluid reasoning (counterfactual letter-string
analogies) at $\rho = +0.55$ Spearman
($n = 82$), and with MMLU-Pro (a proxy for general knowledge) at
$\rho = +0.62$ ($n = 57$). The two correlations are statistically
indistinguishable (Williams' $t$ for the $G_f$-vs-$G_c$ difference
$= +0.04$, $p = 0.96$), mirroring effects reported in human research on intelligence and creativity \citep{gerwig2021intelligence,jauk2013relationship}.

Notably, the same separability holds for model size. Among the 42
open-weight models with disclosed parameter counts, total parameters
correlate with the composite at $\rho = +0.69$, weakening to
$\rho = +0.37$ under active-parameter accounting
(Appendix~\ref{app:aa_battery}). Mixture-of-experts architectures
reach the upper tier on total-parameter rankings without delivering
matched creative ability per active parameter, suggesting model size is a real but
incomplete predictor of \textit{C}. Taken together, \textit{C} tracks general intelligence and model
scale at moderate-to-strong magnitudes, consistent with a creative
ability that draws on both reasoning and knowledge yet is not
reducible to either. 

\subsection{The paired-human subset is more sensitive to creativity-specific prompting than to enabling reasoning}
\label{sec:results_intervention}

A creativity benchmark should respond more to manipulations that
target creativity than to manipulations that target reasoning. We
tested this with two within-model interventions on AGC-Human, run
on separate model cohorts because not all models support a clean
reasoning-mode toggle. Prompting 18 frontier models to ``be
creative'' rather than ``be effective'' produced a large upward
shift ($d_z = +1.40$ length-residualized) that did not sacrifice
quality, with $99.4\%$ of $2{,}700$ generations passing a 3-judge
validity gate. In contrast, enabling reasoning (vs. baseline) on 10 different reasoning-capable models produced a much smaller composite shift
($d_z = +0.34$), roughly one-quarter the magnitude. The benchmark
thus moves more under the manipulation that targets creativity than
under the manipulation that targets reasoning, consistent with the
construct sensitivity expected of a creativity instrument (see Appendix~\ref{app:intervention} for full intervention
designs and per-task results).

\subsection{AGC-Bench predicts which models best agree with human creativity ratings}
\label{sec:muce_results}

For external validation against human-rated data, we asked whether a
model's \benchname{} score predicts its agreement with human creativity
ratings on an independent corpus of 1{,}862 responses to 6 creativity tasks. We find the \benchname{} composite
predicts each model's Pearson correlation (agreement) with human ratings at
$r = +0.66$, $\rho = +0.73$ ($n = 74$ models with AA Intelligence Index coverage), contributing unique variance
beyond the reasoning and knowledge tests (partial $r = +0.34$). Models that score well on \benchname{} thus also rate creativity more like humans do.

\section{Conclusion}
\label{sec:discussion}

\benchname{} provides a psychometric tool for LLM creativity
evaluation, offering insight into the domain-specificity of
creativity and how creative ability relates to intelligence in AI
systems. The pipeline assembles 67 text-only benchmarks under a
PRISMA-compliant review, applies IRT-calibrated LLM-judge training
for automated creativity scoring, and releases the open-weight
AGC-Judge scorer that reproduces the leaderboard at frontier-judge
accuracy. Using factor analysis, we find that verbal creativity is
strongly unidimensional, related to but separable from knowledge and
reasoning, with frontier models competitive with human performance
on ideation and creative writing tasks. We view \benchname{} as
scaffolding for the next phase of work on LLM creativity, with
future efforts targeting shorter scales for cost-efficient evaluation
\citep{polo2024tinybenchmarks,kipnis2025metabench}, extension into
different modalities such as audio and vision, and ultimately the
use of these tools to improve creativity in current models.

\paragraph{Limitations.}
A few caveats are worth noting. The composite creativity score is
specific to this cohort, so scores have no theoretical ceiling
similar to Chatbot Arena ELO---a feature of any benchmark scored
against frontier peers rather than a defect. \benchname{} is currently
English-dominant and limited to image-only inputs. Both extensions
are natural targets for follow-on work, especially the visual tasks,
which are slated for the next release of \benchname{}. Finally,
although our benchmark covers many creative tasks, some aspects of
creativity remain harder to assess, including transformational
creativity, where a model reshapes the conceptual space itself
rather than simply recombining within it
\citep{schapiro2025transformational}. Nevertheless, our
findings provide the first large-scale test of the
structure of artificial creativity and how it may scale with
increasing model capability.

\section*{References}

{
\renewcommand{\section}[2]{}%
\bibliographystyle{plainnat}
\bibliography{references}
}

\appendix
\setcounter{figure}{0}
\setcounter{table}{0}
\renewcommand{\thefigure}{S\arabic{figure}}
\renewcommand{\thetable}{S\arabic{table}}
\renewcommand{\theHfigure}{S\arabic{figure}}
\renewcommand{\theHtable}{S\arabic{table}}

\clearpage

\section{PRISMA flow diagram}
\label{app:prisma}


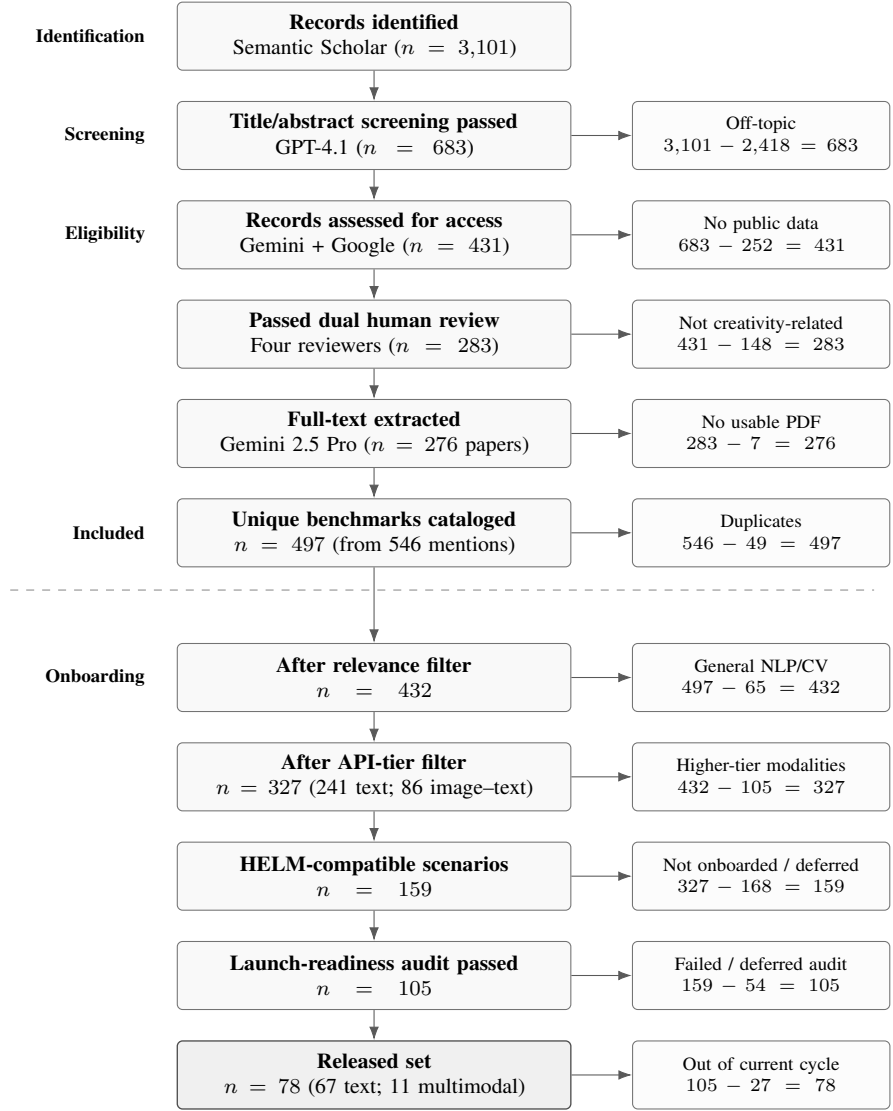
\begin{figure}[!ht]
  \centering
  \begin{tikzpicture}[
    font=\footnotesize,
    node distance=4mm and 8mm,
    box/.style={
      draw=black!55, line width=0.4pt, rounded corners=2pt,
      fill=gray!5, align=center, inner sep=3pt,
      minimum width=52mm, minimum height=9mm, text width=49mm,
    },
    excl/.style={
      draw=black!55, line width=0.4pt, rounded corners=2pt,
      fill=gray!2, align=center, inner sep=2.5pt,
      minimum width=34mm, minimum height=9mm, text width=32mm,
      font=\scriptsize,
    },
    final/.style={
      draw=black!70, line width=0.5pt, rounded corners=2pt,
      fill=gray!12, align=center, inner sep=3pt,
      minimum width=52mm, minimum height=9mm, text width=49mm,
    },
    tier/.style={
      font=\scriptsize\bfseries, anchor=east, align=right, text width=18mm,
    },
    arr/.style={-{Latex[length=1.8mm,width=1.3mm]}, line width=0.4pt, black!65},
    divider/.style={dashed, black!45, line width=0.4pt},
  ]

  \node[box] (ident) {\textbf{Records identified}\\Semantic Scholar ($n=3{,}101$)};
  \node[tier, left=3mm of ident] {Identification};

  \node[box, below=of ident] (screen)
    {\textbf{Title/abstract screening passed}\\GPT-4.1 ($n=683$)};
  \node[tier, left=3mm of screen] {Screening};
  \node[excl, right=of screen] (screen_ex) {Off-topic\\$3{,}101-2{,}418=683$};

  \node[box, below=of screen] (access)
    {\textbf{Records assessed for access}\\Gemini + Google ($n=431$)};
  \node[tier, left=3mm of access] {Eligibility};
  \node[excl, right=of access] (access_ex) {No public data\\$683-252=431$};

  \node[box, below=of access] (review)
    {\textbf{Passed dual human review}\\Four reviewers ($n=283$)};
  \node[excl, right=of review] (review_ex) {Not creativity-related\\$431-148=283$};

  \node[box, below=of review] (extract)
    {\textbf{Full-text extracted}\\Gemini 2.5 Pro ($n=276$ papers)};
  \node[excl, right=of extract] (extract_ex) {No usable PDF\\$283-7=276$};

  \node[box, below=of extract] (catalog)
    {\textbf{Unique benchmarks cataloged}\\$n=497$ (from 546 mentions)};
  \node[tier, left=3mm of catalog] {Included};
  \node[excl, right=of catalog] (catalog_ex) {Duplicates\\$546-49=497$};

  \coordinate (div_left)  at ($(catalog.south west)+(-22mm,-3mm)$);
  \coordinate (div_right) at ($(catalog.south east)+(40mm,-3mm)$);
  \draw[divider] (div_left) -- (div_right);

  \node[box, below=10mm of catalog] (filter)
    {\textbf{After relevance filter}\\$n=432$};
  \node[tier, left=3mm of filter] {Onboarding};
  \node[excl, right=of filter] (filter_ex) {General NLP/CV\\$497-65=432$};

  \node[box, below=of filter] (api)
    {\textbf{After API-tier filter}\\$n=327$ (241 text; 86 image--text)};
  \node[excl, right=of api] (api_ex) {Higher-tier modalities\\$432-105=327$};

  \node[box, below=of api] (helm)
    {\textbf{HELM-compatible scenarios}\\$n=159$};
  \node[excl, right=of helm] (helm_ex) {Not onboarded / deferred\\$327-168=159$};

  \node[box, below=of helm] (audit)
    {\textbf{Launch-readiness audit passed}\\$n=105$};
  \node[excl, right=of audit] (audit_ex) {Failed / deferred audit\\$159-54=105$};

  \node[final, below=of audit] (release)
    {\textbf{Released set}\\$n=78$ (67 text; 11 multimodal)};
  \node[excl, right=of release] (release_ex) {Out of current cycle\\$105-27=78$};

  \draw[arr] (ident)   -- (screen);
  \draw[arr] (screen)  -- (access);
  \draw[arr] (access)  -- (review);
  \draw[arr] (review)  -- (extract);
  \draw[arr] (extract) -- (catalog);
  \draw[arr] (catalog) -- (filter);
  \draw[arr] (filter)  -- (api);
  \draw[arr] (api)     -- (helm);
  \draw[arr] (helm)    -- (audit);
  \draw[arr] (audit)   -- (release);

  \draw[arr] (screen)  -- (screen_ex);
  \draw[arr] (access)  -- (access_ex);
  \draw[arr] (review)  -- (review_ex);
  \draw[arr] (extract) -- (extract_ex);
  \draw[arr] (catalog) -- (catalog_ex);
  \draw[arr] (filter)  -- (filter_ex);
  \draw[arr] (api)     -- (api_ex);
  \draw[arr] (helm)    -- (helm_ex);
  \draw[arr] (audit)   -- (audit_ex);
  \draw[arr] (release) -- (release_ex);

  \end{tikzpicture}
  \caption{PRISMA-based flow of papers and benchmarks through the
  meta-benchmark pipeline. Top five tiers follow PRISMA 2020 reporting
  conventions; the dashed rule separates the systematic review from the
  benchmark onboarding stage, which filters the cataloged benchmarks to
  API-runnable text and image-text tiers and converts the released set
  into runnable evaluation scenarios. Right-hand boxes show the arithmetic for
  the transition into the corresponding left-hand box.}
  \label{fig:prisma}
\end{figure}

\FloatBarrier

\section{Detailed onboarding methodology}
\label{app:onboarding}

This appendix provides the detailed pipeline architecture, failure-mode
analysis, and robustness additions that support the summary onboarding
description in Section~\ref{sec:onboarding}.

\subsection{Agentic pipeline architecture}
\label{app:onboarding:pipeline}

Manual onboarding of 327 candidate benchmarks (241 text-to-text plus
86 image-to-text) would have been prohibitively slow, so we developed
an agentic workflow built on Claude
Code. The workflow is encapsulated in a reusable skill\footnote{A Claude
Code \textit{skill} is a structured set of instructions, code, and
reference materials that directs an agent through a multi-step task.}
(\texttt{benchmark-onboarder}) implemented as a deterministic
nine-state machine that walks each candidate from PDF parsing through
data loading, semantic validation, prompt and judge extraction,
scenario codegen, code review against the paper's stated metric, and a
small-instance pilot run before emit. State transitions are controlled
by deterministic code rather than agent decisions, following design
principles from recent work on production LLM agents, and LLM
reasoning is bounded to specific states with well-defined inputs and
outputs. Each onboarding run used \texttt{google/gemini-2.0-flash} for
structured extraction and \texttt{anthropic/claude-sonnet-4} as the
review and judge model. The full
skill specification, per-state prompts, and transition logic are
released in the artifact bundle (\texttt{skills/benchmark-onboarder/}).

\subsection{Human-in-the-loop verification}
\label{app:onboarding:human}

Although the agentic workflow automated the bulk of scenario construction,
each onboarded benchmark passed through a human review step before
acceptance. Four team members were each assigned 60--71
candidate benchmarks (mean 68) and reviewed the agent's output for prompt fidelity,
dataset-schema correctness, and evaluation-function appropriateness.
Reviewers verified URLs against the original paper, corrected errors in
the scenario file, and either approved the benchmark for inclusion or
documented the reason for rejection in a shared tracking file. A
\texttt{learnings.md} document accumulated cross-team discoveries about
common failure modes and was consulted before each onboarding session;
entries were added automatically by the agent whenever it encountered a
new dataset quirk (e.g., test splits without labels, fields requiring
non-standard loading parameters, field-name mismatches between
documentation and data).

\subsection{Failure modes and mitigations}
\label{app:onboarding:failures}

A review of 155 completed onboarding attempts surfaced four recurring
failure patterns at the agent stage (before human review and
acceptance), each addressed by a specific validation step in the
pipeline. The most common pattern was \textit{hallucinated evaluation
logic}, which arose when the source paper did not pin down a scoring
function precisely; the \textsc{validate-impl} state now compares the
agent's generated metric against the paper's stated metric before
pilot execution, and the agent must cite a source file or paper
section for each evaluation function. The second-most-common pattern
was \textit{missing LLM-judge configuration}, addressed by requiring
judge-field extraction in the \textsc{extract} state. \textit{Prompt
hallucination} (the agent paraphrasing rather than extracting prompts)
was reduced by prioritizing regex and code-based extraction over LLM
paraphrasing and flagging any prompt that could not be traced to a
specific source location. Lower-frequency failures (metric-name
mismatches between paper and code, semantic drift from the paper's
task, multimodal benchmarks silently downcast to text-only,
task-direction inversions) are documented in the released learnings
index alongside their corresponding pipeline checks.

\section{Example prompts by domain}
\label{app:example_prompts}

We illustrate \benchname{}'s source-preserving prompt format with one
example per text-only domain. Each prompt is what the model under
evaluation actually sees, reproduced verbatim from the source paper's
released materials and re-formatted only for HELM's instance schema.
The full per-benchmark prompt templates are released alongside the
harness in the project repository (\projecturl).

\paragraph{Brainstorming \textit{(sdat)}.}
\begin{quote}\small\ttfamily
Please enter 10 words that are as different from each other as possible, in all meanings and uses of the words. Rules: Only single words in English. Only nouns (e.g., things, objects, concepts). No proper nouns (e.g., no specific people or places). No specialised vocabulary (e.g., no technical terms). Think of the words on your own (e.g., do not just look at objects in your surroundings). Make a list of these 10 words, a single word in each entry of the list.
\end{quote}

\paragraph{Problem Solving \textit{(macgyver)}.}
\begin{quote}\small\ttfamily
Your desk drawer has become misaligned and difficult to open. However, the desk is made of metal. Tools available include a bottle of water, a keyring, a pair of pajamas, a hand towel, a coaster, a bottle opener, a shoebox, a pair of scissors, a coffee mug, a makeup compact, a paint roller, an ice cube tray. How do you realign the drawer and open it without damaging its contents?\\
\\
Give a valid (feasible and efficient) solution very concisely. Use step1, step2, etc, and mention the tools to achieve each step. Use as few steps as possible and the answer should ideally be less than 100 words.\ldots
\end{quote}

\paragraph{STEM \textit{(futuregen)}.}
\begin{quote}\small\ttfamily
You are an AI trained to analyze scientific research and suggest future directions based on the content of a paper. Below, you will find sections from a scientific article including the `Abstract', `Introduction', and `Conclusion'. Based on these details, please generate comprehensive and plausible future work suggestions that could extend the research findings, address limitations, and propose new avenues for exploration.\\
\\
\{abstract / introduction / conclusion of the paper\}\ldots
\end{quote}

\paragraph{Story / Narrative \textit{(eqbench\_creative\_writing\_v3)}.}
\begin{quote}\small\ttfamily
Darkroom Revelations\\
Write a first-person narrative from the perspective of Dani, a 25-year-old assertive art student known for her bold style and take-charge attitude in relationships. For the past semester, she's been quietly fascinated by Elliot\ldots Set during one of these intimate darkroom sessions where the red light creates a confessional atmosphere. Explore Dani's complex attraction\ldots The narrative should capture the tender awkwardness of their conversations\ldots
\end{quote}

\paragraph{Figurative Language \textit{(metaphor\_generation)}.}
The scenario passes the literal sentence directly to the model under
the HELM completion template, with no instruction wrapper:
\begin{quote}\small\ttfamily
I wrestled with this decision for years
\end{quote}

\paragraph{Humor \textit{(showerthoughts)}.}
\begin{quote}\small\ttfamily
Please generate one Showerthought, which is inspired by the Reddit community r/Showerthoughts. Try to be clever, creative, and funny. The Showerthought should be relatable and connected to things that people might encounter during mundane tasks. Return only the Showerthought text as one concise standalone statement.\\
\\
Showerthought:
\end{quote}

\section{Per-dataset domain assignments}
\label{app:taxonomy}

Each benchmark maps to one of the \domaincount{} domains in
the creativity taxonomy (Section~\ref{sec:taxonomy}). The assignments
come from a three-model LLM panel (Gemini~3~Flash, Grok-4.1-Fast,
GPT-4.1-mini) prompting on the benchmark's title, task description,
example prompts, scoring method, and source-paper abstract, with
disagreements resolved by majority vote and edge cases adjudicated
by the lead authors. Mean pairwise Cohen's $\kappa \approx 0.85$
across the three LLM raters (Fleiss' $\kappa \approx 0.85$ on the
released $n = 75$ taxonomy panel) supports the domain assignment.
Per-dataset domain assignments, scoring protocol, source paper, license, canonical metric, and item-level statistics are released as
machine-readable CSV.
%

\section{Canonical scoring-metric heterogeneity}
\label{app:metric_types}

Across the \benchmarkcount{} onboarded benchmarks, primary scoring
metrics fall into three families (Table~\ref{tab:metric_types_app}).
The majority are formula-based (exact-match, F1, ROUGE, BLEU,
validity-vote counts, parse rates, and similar source-paper-canonical
formula metrics), a substantial minority use expert-rubric LLM-judge
ratings (Likert scales, typically 1--5 or 1--6, with a smaller subset
on wider 0--100 or 10--50 scales), and a small number use embedding-
or model-based semantic-distance metrics (sentence-BERT, DSI,
contextual entailment).

\begin{table}[h]
\caption{Primary canonical metric category per benchmark across the
\benchmarkcount{} onboarded benchmarks (full per-benchmark mapping
released in the artifact bundle as \texttt{registry\_metrics.yaml}).}
\label{tab:metric_types_app}
\centering
\small
\begin{tabular}{lr}
\toprule
Primary metric category & Benchmarks \\
\midrule
Formula-based (exact-match, F1, ROUGE, BLEU, vote counts, parse rates) & 58 \\
LLM-as-judge (Likert / wide-range rubric ratings) & 19 \\
Model-based semantic-distance (sentence-BERT, DSI, entailment) & 1 \\
\bottomrule
\end{tabular}
\end{table}

\section{Models}
\label{app:models}

The \finalistcount{} models we evaluated span 22 providers and cover
proprietary, open-weight, and reasoning-focused variants released
between 2024 and 2026. Table~\ref{tab:models} reports the
top 30 models ranked by \benchname{} composite. The full
per-model roster (including provider, version string, parameter count where publicly documented, access method, reasoning-mode
support, datasets covered, mean and median $z$, and rank) is released as machine-readable CSV.

\begin{table}[h]
\caption{Top 30 models by JRT-corrected \benchname{} composite
($z$-mean across the \textonlycount{} text-only datasets,
re-z'd within the \finalistcount{} models we evaluated). $n_{\text{ds}}$ is the number of datasets on which the model produced a parseable canonical-metric score within the JRT pipeline. All entries shown are models that scored on at least 65 of 67 datasets.}
\label{tab:models}
\centering\footnotesize
\renewcommand{\arraystretch}{1.05}
\begin{tabular}{rllrrr}
\toprule
\textbf{Rank} & \textbf{Provider} & \textbf{Version} & \textbf{$n_{\text{ds}}$} & \textbf{Mean $z$} & \textbf{Median $z$} \\
\midrule
 1 & anthropic & \texttt{claude-opus-4.7} & 67 & $+0.804$ & $+0.884$ \\
 2 & openai & \texttt{gpt-5.4} & 67 & $+0.737$ & $+0.931$ \\
 3 & openai & \texttt{gpt-5.5} & 67 & $+0.736$ & $+0.796$ \\
 4 & z-ai & \texttt{glm-5.1} & 66 & $+0.711$ & $+0.697$ \\
 5 & z-ai & \texttt{glm-5-turbo} & 66 & $+0.668$ & $+0.783$ \\
 6 & anthropic & \texttt{claude-opus-4.6-fast} & 66 & $+0.663$ & $+0.822$ \\
 7 & z-ai & \texttt{glm-5} & 66 & $+0.651$ & $+0.700$ \\
 8 & openai & \texttt{gpt-5.1} & 66 & $+0.643$ & $+0.790$ \\
 9 & google & \texttt{gemma-4-31b-it} & 67 & $+0.642$ & $+0.670$ \\
10 & moonshotai & \texttt{kimi-k2.5} & 67 & $+0.635$ & $+0.730$ \\
11 & moonshotai & \texttt{kimi-k2-0905} & 66 & $+0.612$ & $+0.655$ \\
12 & z-ai & \texttt{glm-4.7} & 66 & $+0.595$ & $+0.643$ \\
13 & anthropic & \texttt{claude-opus-4.5} & 67 & $+0.585$ & $+0.772$ \\
14 & anthropic & \texttt{claude-opus-4.6} & 65 & $+0.579$ & $+0.837$ \\
15 & google & \texttt{gemma-4-26b-a4b-it} & 67 & $+0.539$ & $+0.610$ \\
16 & google & \texttt{gemini-2.5-flash} & 66 & $+0.508$ & $+0.542$ \\
17 & moonshotai & \texttt{kimi-k2.6} & 67 & $+0.490$ & $+0.618$ \\
18 & anthropic & \texttt{claude-sonnet-4.5} & 67 & $+0.489$ & $+0.767$ \\
19 & anthropic & \texttt{claude-sonnet-4.6} & 67 & $+0.480$ & $+0.707$ \\
20 & deepseek & \texttt{deepseek-v3.2-exp} & 66 & $+0.465$ & $+0.589$ \\
21 & openai & \texttt{gpt-5.4-mini} & 66 & $+0.448$ & $+0.525$ \\
22 & deepseek & \texttt{deepseek-v3.1-terminus} & 66 & $+0.444$ & $+0.581$ \\
23 & openai & \texttt{gpt-4.1} & 66 & $+0.432$ & $+0.414$ \\
24 & deepseek & \texttt{deepseek-v3.2} & 66 & $+0.420$ & $+0.491$ \\
25 & qwen & \texttt{qwen3.5-flash-02-23} & 67 & $+0.388$ & $+0.363$ \\
26 & qwen & \texttt{qwen3-next-80b-a3b-instruct} & 66 & $+0.382$ & $+0.382$ \\
27 & anthropic & \texttt{claude-sonnet-4} & 66 & $+0.381$ & $+0.603$ \\
28 & baidu & \texttt{ernie-4.5-300b-a47b} & 66 & $+0.379$ & $+0.321$ \\
29 & mistralai & \texttt{mistral-medium-3.1} & 67 & $+0.357$ & $+0.399$ \\
30 & deepseek & \texttt{deepseek-chat-v3.1} & 66 & $+0.325$ & $+0.407$ \\
\bottomrule
\end{tabular}
\end{table}


Additional model providers (\texttt{ai21}, \texttt{allenai}, \texttt{arcee-ai}, \texttt{deepcogito}, \texttt{ibm-granite}, \texttt{minimax}, \texttt{morph}, \texttt{nex-agi}, \texttt{nousresearch}, \texttt{tencent}) each contribute one to four models. The released roster CSV documents per-provider model counts for them alongside \texttt{nvidia}, \texttt{meta-llama}, \texttt{amazon}, \texttt{baidu}, \texttt{mistralai}, \texttt{deepseek}, \texttt{moonshotai}, \texttt{qwen}, \texttt{google}, \texttt{z-ai}, \texttt{openai}, and \texttt{anthropic}.

\section{Data-quality audit details}
\label{app:dq}


The audit summarized in Section~\ref{sec:dq} validates that every model produces on-task, non-degenerate output on every dataset.

\paragraph{Judge model and rubric.}
The audit judge was \texttt{x-ai/grok-4.1-fast}, chosen for parser
robustness against verbose judge replies. The audit measures
\textit{presence and legibility} (binary \texttt{on\_task} and
\texttt{garbled} flags) rather than rubric scoring, so reuse of a
JRT-panel judge here does not validate the panel's own Likert
verdicts; the cross-family validation pass below (using a
non-JRT-panel second judge) addresses that question separately. The
judge sampled three random instances per cell, saw the rendered
prompt and the full completion, and returned three boolean flags per
response: \textit{on-task}, \textit{garbled} (empty, refusal,
gibberish, or cut off before producing meaningful content), and
\textit{uninterpretable} (rare). The cell-level composite is
$\texttt{valid\_response} = \neg \texttt{garbled} \wedge \texttt{on\_task}$.
Multimodal cells were re-rated by the same judge under a vision-aware
pass with the input image attached.

\paragraph{Coverage.}
%
The audit covered $6{,}635$ text-only model-dataset combinations across the full set of evaluated models (a superset of the \finalistcount{} used in the main analyses), with zero judge errors.
The on-task rate was 95.1\% and the invalid rate was 5.5\%. We additionally ran a vision-aware pass on the 730 multimodal model-dataset combinations, obtaining 94.4\% on-task rate and 6.1\% invalid rate.

\paragraph{Validation.}
A 2{,}725-cell stratified subsample was re-rated by
\texttt{deepseek-v4-flash}, drawn from a different model family than
the primary audit judge and outside the JRT scoring panel. The binary
\texttt{valid\_response} composites agreed at $\kappa = 0.67$
(substantial agreement on the Landis--Koch scale). Per-instance
Likert agreement was not computed for this overlap because the v4
audit pass recorded only the binary validity flags rather than a
1--5 quality score.

\paragraph{Truncation patches.}
Frontier 2026 models produce verbose outputs that exceeded the original
source-paper token budgets on some open-ended generation datasets. We
detected truncation by comparing generation length to the configured
\texttt{max\_tokens} and re-ran 16 datasets at raised token caps
(typically $4{,}096 \to 8{,}192$). \texttt{critics\_story} still exceeded
budget at the higher cap and was excluded. \texttt{fann\_or\_flop} was retained despite verbose-tail truncation, with the per-dataset truncation rate documented
as a caveat (Section~\ref{sec:onboarding}).

\paragraph{Pre-release data-quality sweep and cascade impact.}
\label{app:dq_sweep}
A pre-release sweep cross-referenced the on-task audit above with
heuristic flags on the raw \texttt{scenario\_state} files (empty
output, refusal, repetitive sequence collapse). 32 cells flagged by
both signals were masked at the score level (\texttt{dq\_masked} in
the released long table) rather than scored as failed attempts.
Cascade-impact analysis confirmed the c-factor (eigenvalue 4.89,
$\alpha = 0.96$, 81.5\% variance) and the intelligence convergent-
validity correlations (LSA, GPQA-Diamond, AA Intelligence Index) are
unchanged to two decimals under the mask versus an unmasked run, so
the mask removes degenerate cells without altering substantive
findings. Per-cell flags and the cascade-impact log are released in
the artifact bundle (\texttt{audit/dq\_sweep/}).

\section{Judge calibration: GRM specification, audit, and AGC-Judge details}
\label{app:jrt}

This appendix specifies the planned-missing 2-of-3 design (PM1-50), the
Bayesian Graded Response Model used to fit per-judge severity and
per-unit ability, the parser audit that recovered silent-zero
contamination from earlier annotators, the per-benchmark inter-judge
agreement table, and the AGC-Judge fine-tuning details summarized in
Section~\ref{sec:jrt}.

\subsection{Planned-missing 2-of-3 design and parser audit}
\label{app:jrt:pm1}

Each of the $92{,}080$ (benchmark, model, item) units in the $24$
LLM-judge canonical-metric cells was rated by two of three vendor
judges (\texttt{google/gemini-3-flash-preview},
\texttt{x-ai/grok-4.1-fast}, \texttt{openai/gpt-4.1-mini}) under a
planned-missing 2-of-3 design, with each unit assigned one of three
judge pairs (seed 42). The design targets $184{,}160$
(rater $\times$ unit) cells; $182{,}924$ produced a non-sentinel
rating ($99.3\%$ completion) and the residual missing observations
are masked at the unit level in the joint-rater fit. A
data-integrity patch resolved a pre-existing first-integer parser in
$23$ of $35$ LLM-judge annotators that returned silent zeros on parse
failures and grabbed stray integers from verbose judge replies. The
patched parser prefers labeled patterns (``Score: $N$''), falls back
to last-integer-in-range, and emits the missing-data sentinel
otherwise; it reduced contamination from $0.32\%$ to $0\%$, with
historical out-of-range ratings re-coded as missing for the JRT fit.

\subsection{Bayesian Graded Response Model}
\label{app:jrt:grm}

For each (benchmark, metric) cell with $N$ units and $R$ raters, we
fit per-unit ability $\theta_p$, per-rater discrimination $\alpha_r$,
and per-rater ordered thresholds $\beta_{r, k}$ via a Graded Response
Model:
\begin{equation}
P\!\left(\text{rating}_{r, p} \geq k+1\right) = \sigma\!\left(\alpha_r (\theta_p - \beta_{r, k})\right),
\end{equation}
with $\theta_p \sim \mathcal{N}(0, 1)$, $\log \alpha_r \sim
\mathcal{N}(0, 0.3)$, and ordered $\beta_{r, :}$ drawn from
$\mathcal{N}(0, 2)$. Inference uses NumPyro stochastic variational
inference (AutoNormal guide, Adam at learning rate $0.05$, $800$
update steps per cell), fit cell-by-cell so per-cell rater identifiers
and rubric scales remain isolated. All $24$ cells reached converged
ELBO; per-judge severity ranges from $\beta = -0.90$
(\texttt{gpt-4.1-mini}) to $\beta = +0.04$ (\texttt{gemini}), with
\texttt{grok-4.1-fast} at $\beta = -0.46$. Full pipeline configuration
and per-cell fit diagnostics are released in the artifact bundle.

\subsection{Per-benchmark inter-judge agreement}
\label{app:jrt:per_bench}

Pairwise Spearman $\rho$ on co-rated units spans $0.30$
(\texttt{showerthoughts}) to $0.88$
(\texttt{eqbench\_creative\_writing\_v3}); agreement is highest for
benchmarks with concrete rubrics and lowest for short, subjective
tasks. The pattern is rater-invariant (all three judge
pairs track together across benchmarks), so low agreement reflects
task ambiguity rather than a single outlier judge. The full
per-(benchmark, judge-pair) table is released as machine-readable CSV
in the artifact bundle.

\subsection{Drift of validation correlations under JRT correction}
\label{app:jrt:drift}

The JRT correction recalibrates per-unit ability $\theta$ without
destroying the rank signal raw means already carry. Per-unit raw and
JRT-corrected scores correlate at $\rho = 0.97$, and the 
composite leaderboard recomputed against JRT-corrected supervision
agrees with the single-judge leaderboard at Spearman $\rho = 0.95$.
The c-factor itself is preserved (eigenvalue $4.89$, $\alpha = 0.96$,
loadings $+0.87$ to $+0.94$ across the six domains), and
convergent validity against intelligence indicators tightens slightly
under the corrected supervision (LSA Pearson $r = +0.51 \to +0.53$;
AA Intelligence Index Spearman $\rho = +0.67 \to +0.69$, with all
five AA indicators moving in the same direction). The correction
therefore removes the per-judge severity offsets reported in
Section~\ref{sec:jrt} without destabilizing the substantive findings.

\subsection{AGC-Judge training details}
\label{app:jrt:agcjudge}

\paragraph{Theta-to-integer SFT target.}
The GRM produces a continuous latent ability $\theta_p$ on the standard
normal scale, but AGC-Judge fine-tuning targets the integer rubric range
each source paper uses (1--5, 10--50, 0--100, etc.). For each unit the
training target is recovered by inverting the GRM at the per-rater
severity thresholds: the integer score $k \in \{1, \ldots, K\}$ chosen
is the one for which the GRM's cumulative probability
$P(\text{rating}_{r,p} \geq k)$ crosses $0.5$ under the calibrated
$(\alpha_r, \beta_{r,k})$ for the rater whose severity is closest to
zero (the cohort-canonical reference rater), then rounded to the
nearest integer in the rubric range. The resulting target is on the
source paper's native scale and is what AGC-Judge predicts directly.

Training data was filtered to drop the 4.5\% of rows whose total
chat-template length exceeded 7{,}500 approximate tokens
(characters $/$ 4), to fit within the Together fine-tune harness's
8{,}192-token sequence budget while leaving headroom for tokenizer
drift. The filter dropped 2{,}298 of 50{,}597 training rows and 345 of
7{,}228 validation rows. Token outliers concentrated in benchmarks
with long instruction text (Arabic-language story benchmarks,
multi-turn dialogue benchmarks) and contribute negligible learning
signal at high cost in compute. The held-out test, held-out-models,
and held-out-benchmarks splits retain outliers to evaluate
generalization across the full input distribution. Other
hyperparameters: LoRA $r = 16$, $\alpha = 32$, learning rate
$1\mathrm{e}{-4}$, 1 epoch, AdamW with default Together SFT settings
(\texttt{train\_on\_inputs = "auto"}, masking system and user tokens
during loss computation). The fine-tune ran in 20 minutes on
$2 \times$ H100-80GB SXM at a one-time cost of \$211. The released held-out evaluation reports parse-failure rates and
per-bench correlations on the full splits (test $N = 14{,}457$,
held-out-models $N = 10{,}966$, held-out-benchmarks $N = 8{,}803$),
with the full held-out-bench scatter (pooled $\rho = 0.83$) released
in the artifact bundle. AGC-Judge is trained exclusively on LLM-produced responses
with frontier-LLM ratings; for cross-source (human vs.\ LLM) scoring
on AGC-Human cells, we use a fairness-aware inference prompt that
discloses the LLM self-preference bias documented in prior judge work
\citep{panickssery2024llms, ye2024justice}, instructs the judge to
focus on the underlying idea rather than stylistic register, and
labels the response source before scoring (an instance of the
prompt-based debiasing strategies validated by
\citealp{liu2025agdejudge} and \citealp{wu2025instajudge}). The
prompt template, validation results, and a counterfactual style-flip
experiment are documented in
Appendix~\ref{app:agc_human_judge}.

\subsection{External validation on the Orwig creative writing data}
\label{app:jrt:orwig}

As an out-of-distribution check that does not appear in any AGC-Bench
training or evaluation split, we ran AGC-Judge on the Orwig short-story
corpus \citep{orwig2024language}: 718 stories on six 3-word cue
triplets, with conditions split as human ($n = 300$), GPT-3
($n = 298$), and GPT-4 ($n = 120$). Each story carries a human-rater
creativity rating averaged across multiple raters, an existing
GPT-judge creativity rating from the original Orwig pipeline, a
divergent-semantic-integration (DSI) score, word count, and a
perceptual-details measure. Critically for the source-bias question,
the Orwig human raters do not show a strong source preference (humans
2.93 vs.\ GPT-4 2.94 mean creativity), so the data offer a fairer test
of AGC-Judge alignment with humans than corpora in which raters
strongly favored one source.

AGC-Judge under the human-source-attribution prompt
(Section~\ref{sec:jrt}) recovered Spearman $\rho = +0.65$ against the
human creativity gold and $\rho = +0.80$ against the existing
LLM-judge gold ($n = 718$). The two correlations were computed on the
same data with the shared variable AGC-Judge, so we tested their
difference with Williams' modification of Hotelling's $t$-test for
dependent correlations: $t = -9.42$, $p < 0.0001$ on Pearson, $t =
-7.98$, $p < 0.0001$ on Spearman. AGC-Judge is significantly more
aligned with the LLM-judge gold than with the human gold. Two pieces
of context matter for interpreting that difference. First, AGC-Judge's
correlation with human creativity ($\rho = +0.65$) is essentially
identical to the original GPT-judge's correlation with humans on the
same data ($\rho = +0.66$): AGC-Judge predicts human ratings about as
well as a frontier GPT judge does, while also tracking the LLM-judge
construct more tightly. Second, the source-preference gap on AGC-Judge
(GPT-4 vs.\ human mean) is $+0.51$ on the 1--5 scale, compared with
$+0.83$ for the original GPT-judge and $+0.01$ for human raters
(Table~\ref{tab:orwig_source}). AGC-Judge inherits some of the
LLM-stylistic preference of its supervision but at roughly 60\% of the
original LLM-judge magnitude.

\begin{table}[h]
\caption{Per-condition mean creativity rating on the Orwig 718-story
corpus, scored by AGC-Judge (this work, JRT-corrected supervision),
human raters (Orwig original), and the GPT-judge in the original Orwig
pipeline. Source-preference gap is the GPT-4 vs.\ human-condition
mean. Smaller gap is better for source neutrality.}
\label{tab:orwig_source}
\centering\small
\begin{tabular}{lrrrr}
\toprule
\textbf{Condition} & \textbf{$n$} & \textbf{AGC-Judge} & \textbf{Human raters} & \textbf{GPT-judge} \\
\midrule
human  & 300 & 2.47 & 2.93 & 3.09 \\
GPT-3  & 298 & 2.38 & 2.85 & 2.95 \\
GPT-4  & 120 & 2.98 & 2.94 & 3.92 \\
\midrule
GPT-4 $-$ human gap & --- & $+0.51$ & $+0.01$ & $+0.83$ \\
\bottomrule
\end{tabular}
\end{table}

Within-condition correlations decompose the alignment further. On
human-written stories, AGC-Judge is equally aligned with human and
LLM-judge gold (both $\rho \approx +0.74$). On LLM-written stories,
AGC-Judge tracks the LLM-judge much more tightly than the human raters
(GPT-4 stories: $\rho = +0.86$ vs.\ $\rho = +0.65$). We treat the
Orwig result as evidence that AGC-Judge generalizes its scoring
construct beyond the AGC-Bench training distribution while inheriting
a moderate, characterizable LLM-stylistic preference from its
supervision.

\section{Intelligence assessments, human-data convergent validity, and intervention designs}
\label{app:intervention}

This appendix details the four external designs summarized in
Section~\ref{sec:eval}: a targeted fluid-reasoning indicator
(letter-string analogies), an external convergent-validity test on
human-rated creativity (MuCE), a be-creative versus be-effective
prompt manipulation, and a reasoning on/off intervention.

\paragraph{Letter-string analogies as a targeted $G_f$ indicator.}
\label{app:lsa}
The human creativity literature situates creative ability primarily
against \emph{fluid} intelligence ($G_f$), the component most directly
implicated in creative idea generation \citep{mcgrew2009chc,
gerwig2021intelligence, stevenson2021minimal}. Standard LLM
intelligence batteries (GPQA, HLE, LiveCodeBench, SciCode, and the
Artificial Analysis composite) blend $G_f$ with substantial crystallized
($G_c$) and quantitative-knowledge ($G_q$) content, making them poorly
suited as a clean creativity--$G_f$ test. We therefore use the
Lewis--Mitchell counterfactual letter-string analogy task
\citep{lewis2024evaluating} as our primary $G_f$ indicator: a
Hofstadter-style rule-abstraction probe whose alphabets are permuted to
block surface-pattern shortcuts. We administer 500 items spanning the
task's transformation types, with reasoning off and temperature zero.
As a robustness check, we also evaluate the cohort on each component of
the AA battery (GPQA, HLE, LiveCodeBench, SciCode, and the AA
composite); the AA-battery comparison appears in
Appendix~\ref{app:aa_battery}.

\paragraph{Convergent validity on independent human creativity ratings (MuCE).}
\label{app:muce}
To test whether \benchname{} tracks creativity in a way that generalizes
beyond our scoring pipeline, we predict human creativity ratings on the
Multitask Creativity Evaluation (MuCE) benchmark
\citep{ismayilzada2025crpo}, an independent collection of human-rated
creative responses spanning more than 30 source studies. We
build a balanced 1{,}862-item subset across 25 source studies and 6 task
families that do not overlap with the paired-human CAP subset
(Real-Life Creative Problem Solving, Consequences, Malevolent Problems,
Question Asking, Essays, Poems). For each model, we ask it to rate every response on the 10--50 integer scale that MuCE uses,
with a prompt invoking the same three-criterion creativity definition
(novel, high-quality, surprising) and conditioned per item on the
human-rater target dimension (originality vs.~creativity). For each
model we compute the Pearson correlation between its predicted ratings
and the human ratings across the 1{,}862-item subset, and the key 
question is whether the \benchname{} composite predicts a model's
agreement with humans across the cohort.

\paragraph{Be-creative versus be-effective intervention.}
\label{app:becreative}
A be-creative prompt manipulation, robustly known in the human
creativity literature to increase originality and diversity on
divergent-thinking tasks (meta-analyses by
\citet{acar2020divergent} and \citet{saidmetwaly2020testing}), was
applied to 18 frontier-cohort models on
the five CAP tasks (AUT, Design, SCTT, Story, Metaphor---see
Section~\ref{sec:cap_human_ai}) with three repetitions per cell. Each
model was prompted twice for every item, once with a be-creative
instruction and once with a be-effective instruction. Composite shifts
are reported using a within-model paired-difference effect size ($d_z$)
on the CrPO composite, with and without length-residualization to
control for the verbosity confound.

\paragraph{Reasoning on/off intervention.}
\label{app:reasoning}
A second intervention toggled provider-supported thinking-mode prompts
on a matched-pairs subset of 10 reasoning-capable frontier systems
(\texttt{deepseek-v3.1-terminus}, \texttt{deepseek-v3.2},
\texttt{gemini-2.5-flash}, \texttt{gemini-3.1-flash-lite-preview},
\texttt{gpt-5.4-mini}, \texttt{qwen3-30b-a3b}, \texttt{grok-4.20},
\texttt{glm-4.5-air}, \texttt{glm-5-turbo}, \texttt{glm-5.1}). Anthropic
Claude variants are excluded because the OpenRouter reasoning-disable
flag was not reliably honoured at the time of the run. Composite shifts
and per-sub-metric (diversity, DSI; surprise was dropped from the
canonical CrPO panel after the alignment-LM bias diagnostic in
Section~\ref{sec:cap_human_ai}) shifts are reported as within-model
paired-difference effect sizes ($d_z$) to characterize whether
reasoning produces uniform gains or splits across sub-metrics.

\section{C-factor robustness}
\label{app:cfactor_robustness}

The c-factor reported in Section~\ref{sec:cfactor}
is a 1-factor
extraction on the $\finalistcount \times \domaincountnr$ per-(model, domain)
composite matrix (eigenvalue $4.89$, $\alpha = 0.96$, $81.5\%$ variance,
parallel-analysis $p_{95} = 1.53$ on $1{,}000$ random matrices). Variance
percentages throughout this appendix follow the eigenvalue-divided-by-domain-count
convention used in the body. We tested three threats to the result:
capability confounding (whether c is a re-parameterization of general
capability), scoring-class artifact (whether c reflects one scoring family
rather than creative ability), and composition sensitivity (whether
the structure depends on the specific set of 83 models). All checks
recompute on the released data and are reproduced in
\texttt{reproduce\_appendix.sh}.

\paragraph{Capability control.}
Counterfactual letter-string analogy accuracy
(Appendix~\ref{app:intervention}) was regressed out of each domain
composite and a 1-factor extraction was performed on the residuals.
Result: eigenvalue $4.61$, $\alpha = 0.94$, $76.8\%$ variance ($n = 82$;
one model lacks LSA coverage). The factor is preserved after
partialling out fluid reasoning, indicating that c is not a
re-parameterization of $G_f$.

\paragraph{Scoring-class invariance.}
We split the benchmarks by primary canonical metric type and
re-extracted a 1-factor solution on the per-(model, domain) composite
restricted to each scoring class. The formula-based subset (40 datasets,
five domains) yielded eigenvalue $3.68$, $\alpha = 0.91$, $73.6\%$
variance. The LLM-judge subset (23 datasets, five domains) yielded
eigenvalue $4.28$, $\alpha = 0.96$, $85.5\%$ variance. The model-based
subset reduces to a single dataset and
is omitted. The two scoring classes that recover meaningful subsets each
emerge as unidimensional, indicating that c is not an artifact of any
single scoring family.

\paragraph{Composition sensitivity.}
Two checks probe the c-factor's dependence on the specific composition.
(a) \emph{Dropping a domain.} Story / Narrative is the largest domain at 20 datasets. We re-ran the factor analysis on the remaining five domains and recovered a similar structure: eigenvalue $4.07$, $\alpha = 0.94$, $81.4\%$ variance. Each remaining domain's variance share barely moves, indicating that the c-factor is not disproportionately driven by Story / Narrative.
(b) \emph{Subsampling models.} We bootstrap-subsampled the 83 models to $k = 60$ across 500 draws (seed 42) and re-extracted the factor each time. Eigenvalue, $\alpha$, and variance all stay tight: eigenvalue $4.92$ ($95\%$ CI $[4.74, 5.05]$), $\alpha = 0.96$ $[0.95, 0.96]$, variance $82.0\%$ $[79.0, 84.2]$. The factor is stable under both perturbations.

The c-factor structure is preserved across all of the above robustness checks.

\section{Artificial Analysis intelligence battery: robustness check}
\label{app:aa_battery}

The Artificial Analysis (AA) intelligence battery aggregates public
benchmarks across knowledge, reasoning, coding, and agentic tasks. We
report it here as a secondary check on the relationship between
\benchname{} and general capability. Our discriminant-validity
analysis uses the pure-$G_f$ letter-string analogy task instead
(Appendix~\ref{app:intervention}) because the AA battery blends fluid reasoning
with substantial $G_q$ and $G_c$ content, has saturated in the top tail,
and covers only a subset of the cohort.

Table~\ref{tab:aa_battery} reports per-indicator Spearman and Pearson
correlations between the JRT-corrected \benchname{} composite and each
AA indicator, restricted to the 
\finalistcount{} models we evaluated
(intersected with AA coverage per indicator).
Correlations range from $\rho = +0.53$ (AA Math Index) to $\rho = +0.85$
(MMMU-Pro), with the AA composite Intelligence Index at $\rho = +0.77$.
The pattern is broadly consistent with the LSA result ($r = +0.53$,
Section~\ref{sec:c_intelligence}) at the lower bound. Coding and
visual-reasoning indicators (Coding Index, MMMU-Pro) sit at the high
end while pure-math indicators (Math Index, AIME) sit at the lower end,
suggesting that creative performance tracks coding, reasoning, and
multimodal knowledge more closely than narrow mathematical-competition
content. The JRT-corrected composite strengthens every correlation by
$\Delta\rho \in [+0.04, +0.13]$ relative to the raw single-judge
composite.

\begin{table}[h]
\caption{Correlations between the JRT-corrected \benchname{} composite
(Section~\ref{sec:jrt}) and indicators on the Artificial Analysis
intelligence battery (AA snapshot 2026-04), on the
\finalistcount{} models we evaluated, 
intersected with
AA coverage per indicator. $n$ varies because not every model has
AA coverage on every indicator.}
\label{tab:aa_battery}
\centering\small
\renewcommand{\arraystretch}{1.15}
\begin{tabular}{lrrr}
\toprule
\textbf{AA indicator} & $n$ & $\rho$ & $r$ \\
\midrule
AA Intelligence Index (composite) & 74 & $+0.769$ & $+0.669$ \\
GPQA-Diamond                      & 75 & $+0.760$ & $+0.651$ \\
Humanity's Last Exam              & 75 & $+0.637$ & $+0.577$ \\
LiveCodeBench (AA)                & 57 & $+0.536$ & $+0.440$ \\
SciCode                           & 75 & $+0.756$ & $+0.673$ \\
AA Coding Index (composite)       & 69 & $+0.775$ & $+0.675$ \\
AA Agentic Index (composite)      & 67 & $+0.745$ & $+0.662$ \\
MMLU-Pro                          & 57 & $+0.620$ & $+0.526$ \\
AA Math Index (composite)         & 53 & $+0.528$ & $+0.406$ \\
MATH-500                          & 33 & $+0.597$ & $+0.567$ \\
MMMU-Pro                          & 35 & $+0.854$ & $+0.776$ \\
$\tau^2$-Bench                    & 70 & $+0.695$ & $+0.594$ \\
IFBench                           & 72 & $+0.667$ & $+0.526$ \\
CritPT                            & 68 & $+0.674$ & $+0.422$ \\
\bottomrule
\end{tabular}
\end{table}

\begin{figure}[h]
\centering
\includegraphics[width=0.92\linewidth]{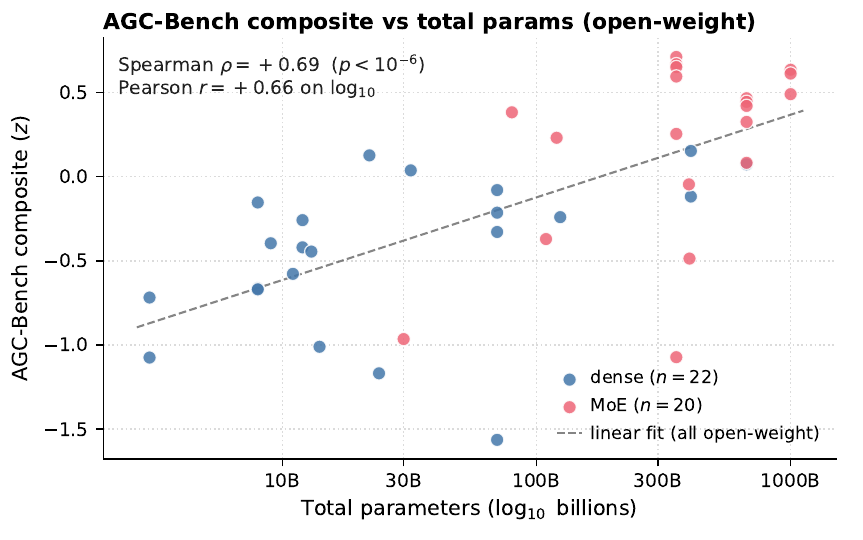}
\caption{\benchname{} composite vs total model parameters on the
$42$ open-weight models with documented parameter counts.
Each point is one model, color-coded by architecture (dense vs.\ MoE).
Closed-frontier models are excluded because providers do not disclose
parameter counts. Spearman $\rho = +0.69$ ($p < 10^{-6}$), Pearson
$r = +0.66$ on $\log_{10}$ total parameters; the same relationship
under active-parameter accounting weakens to $\rho = +0.37$, indicating
that MoE total size reaches the upper tier without delivering matched
creative ability per active parameter.}
\label{fig:param_count_scatter}
\end{figure}

\paragraph{Parameter-count correlation on the open-weight subset.}
Parameter counts are not disclosed for the proprietary frontier
($41$ of $\finalistcount$ models), so a size-vs-creativity
analysis is feasible only on the open-weight subset. Across the
$42$ open-weight models with documented parameter counts,
total parameters correlate with the JRT-corrected \benchname{}
composite at Spearman $\rho = +0.69$ ($p < 10^{-6}$, Pearson
$r = +0.66$ on $\log_{10}$ total parameters,
Figure~\ref{fig:param_count_scatter}). The relationship weakens
substantially under \emph{active}-parameter accounting ($\rho =
+0.37$, $p = 0.015$): mixture-of-experts architectures reach the
upper tier on total-parameter ranking but trail their dense
counterparts when scoring is normalized by active compute. Within
the dense-only sub-cohort ($n = 22$) the correlation is $\rho =
+0.54$ ($p = 0.009$); within the MoE-only sub-cohort ($n = 20$)
total params yield $\rho = +0.33$ ($p = 0.16$, n.s.) and active
params yield $\rho = +0.18$ ($p = 0.45$, n.s.). Total size
therefore predicts creative composite about as well as the AA
Intelligence Index ($\rho = +0.77$), with active-parameter scaling
offering a meaningfully weaker signal. Per-model parameter
assignments and architectural classification are released as
machine-readable CSV in the artifact bundle
(\texttt{model\_size\_correlation.csv}).

\section{AGC-Human rank dissociations and convergent validity with AGC-Judge}
\label{app:agc_human_judge}

\begin{figure}[!htbp]
\centering
\includegraphics[width=0.85\linewidth]{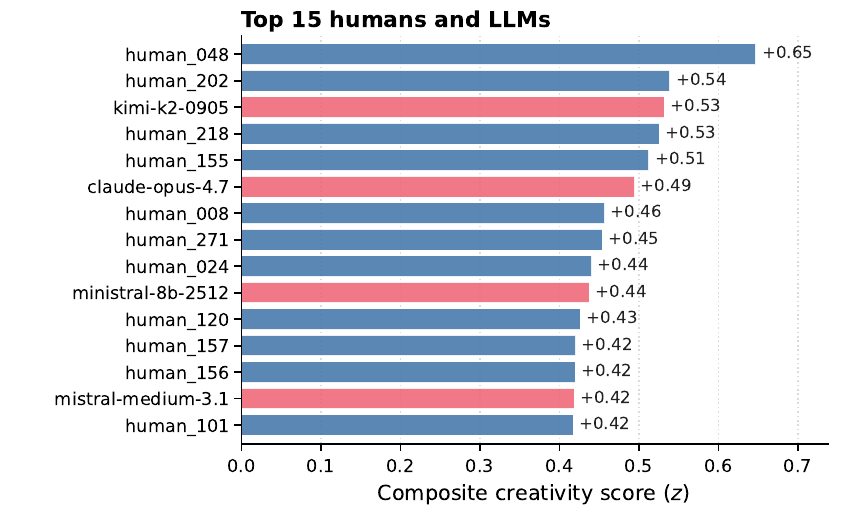}
\caption{Top 15 humans and LLMs by composite creativity score (CrPO
panel, length-residualized $z$), drawn from $201$ humans and $80$
LLMs in the paired-task cohort. The top human (\texttt{human\_048})
leads the top LLM (\texttt{moonshotai/kimi-k2-0905}) by $0.12$ z
($+0.65$ vs.\ $+0.53$).}
\label{fig:human_vs_llm}
\end{figure}

AGC-Human and the broader \benchname{} composite ask the same
question with different instruments: one applies the CrPO panel to
five paired-human CAP tasks, the other aggregates source-paper
canonical metrics across \textonlycount{} datasets. Across the
$80$ LLMs with coverage on both, the two rankings agree at
Spearman $\rho = +0.56$ ($p < 10^{-7}$). The top human reaches a
composite of $+0.65$ and the top LLM
(\texttt{moonshotai/kimi-k2-0905}) reaches $+0.53$, with the human
ahead by $0.12$ z on the canonical two-component panel
(Figure~\ref{fig:human_vs_llm}). The
within-instrument $\alpha$ gap (LLM $0.64$ vs.\ human $0.42$,
$p = 0.028$) speaks to the domain-generality of LLM creativity
rather than to human upper-tail dominance per se: the human-to-LLM
share among the top 30 entities tracks the underlying base rate
($201$ humans of $281$ total $=$ $71.5\%$), so a count of human-led
top ranks adds little beyond the top-of-leaderboard separation
reported above. The CAP subset privileges high-novelty ideational
generation, which diverges from the broader competence captured by
the full \benchname{} cohort.

\paragraph{Per-model rank dissociations.}
\benchname{}-vs-AGC-Human rank shifts on the joint-coverage subcohort
($n = 80$ LLMs scored on both instruments; ranks recomputed within
the intersection) highlight which models the divergent-generation
panel rewards or penalizes. Largest risers:
\texttt{mistralai/ministral-8b-2512} (\#69 $\to$ \#3, $+66$
positions), \texttt{z-ai/glm-4.5v} (\#76 $\to$ \#19), and
\texttt{qwen/qwen3-14b} (\#80 $\to$ \#38). Largest fallers:
\texttt{google/gemini-2.5-flash} (\#14 $\to$ \#80),
\texttt{moonshotai/kimi-k2.5} (\#18 $\to$ \#75), and
\texttt{openai/gpt-4o} (\#22 $\to$ \#59). The pattern is consistent
with the broader cross-instrument correlation ($\rho = 0.56$,
JRT-corrected): agreement is strong on average but heterogeneous at
the model level, with frontier-tier conversational models tending to
underperform on the human-judged composite relative to their
\benchname{} position and several open-weight mid-tier models
outperforming their \benchname{} rank. Three \benchname{}-side models
(\texttt{minimax/minimax-m2-her}, \texttt{morph/morph-v3-fast},
\texttt{nvidia/nemotron-3-nano-30b-a3b}) lack v5-2c coverage and are
excluded from this comparison.

\paragraph{Convergent validity with AGC-Judge.}

To externally validate the AGC-Human composite, we re-scored every cell 
with the open-weight AGC-Judge
(Section~\ref{sec:jrt}). For each (entity, task, prompt) we sent the
response to AGC-Judge with a task-specific creativity
rubric (10--50 scale: ``A creative response is novel and effective''). We
ran two prompt variants on the same 7{,}647 cells:

\begin{itemize}\itemsep1pt\parskip0pt
\item \emph{Standard}: the canonical scoring template used at training time
(rubric, prompt, response, request a single integer).
\item \emph{Fairness-aware}: a counter-bias prompt that discloses the
LLM-self-preference bias documented in prior judge work
\citep{panickssery2024llms, ye2024justice}, instructs the judge to focus on
the underlying idea rather than stylistic register, and discloses the
source (human study participant vs.\ AI language model) before scoring.
\end{itemize}

The standard prompt reproduces a strong LLM-favoring effect: LLMs
averaged 33.5/50 vs.\ humans 22.1/50 (gap $+11.4$, roughly $1.4$~SD).
This is the expected direction and magnitude given that
AGC-Judge's fine-tuning corpus contains exclusively LLM
responses paired with frontier-LLM ratings (48{,}299 rows of
JRT-corrected supervision spanning 24 canonical-metric cells; no
human responses and no human ratings appear in training). The judge
is therefore meeting human-produced creative responses for the first
time at evaluation, which makes a debiasing step both expected and
methodologically necessary for any cross-group claim that uses the
trained judge.
The fairness-aware prompt reduces this gap by $\approx 35\%$ to $+7.4$
(humans 24.1, LLMs 31.5) while preserving cell-level rank ordering against
the standard prompt at $r = +0.91$ and slightly improving convergent
validity with the formal AGC-Human composite ($\rho = +0.53 \to +0.60$).
This is consistent with prior findings that prompt-based debiasing
attenuates self-preference bias without eliminating it
\citep{liu2025agdejudge, wu2025instajudge}.

To test whether the residual bias acts on stylistic register or on
content, we ran a counterfactual style-flip experiment on a sample of
50 cells (5 humans + 5 LLMs per task at the anchor prompt). For each cell
we generated a paraphrase that preserved every idea verbatim while
shifting register --- humans rewritten in polished LLM-style, LLMs
rewritten in casual human-style --- via Grok-4.1-fast (following the
adversarial paraphrase paradigm of \citealp{krishna2023paraphrasing}).
We then re-scored the transformed cell with the standard prompt. Mean
score deltas were near zero in both directions
($\Delta_{\text{polish}} = +0.40$, $d = 0.29$;
$\Delta_{\text{humanize}} = +0.52$, $d = 0.15$ on the 10--50 scale),
indicating that holding the underlying idea constant and swapping
stylistic register barely shifts AGC-Judge. The cross-group bias
under the standard prompt is therefore consistent with content-level
differences in what humans and LLMs produce rather than purely
stylistic-register effects, though the 50-cell probe cannot rule out
register entirely. The fairness-aware prompt's effect appears to
operate by changing how the judge weights ideas more than by changing
how it reads them. The
AGC-Judge fairness-aware composite correlates with the formal
AGC-Human composite at $\rho = +0.60$ (LLMs only: $\rho = +0.76$;
humans only: $\rho = +0.58$), establishing a second instrument that
agrees within group on creative-cell rank ordering while diverging across
groups in the manner expected from a frontier judge with residual
self-preference.

\section{AGC-Human qualitative samples}
\label{app:agc_human_qualitative}

To convey the flavor of top-tier creative responses, we reproduce
verbatim responses to the Alternate Uses Task (BRICK) from the two
top-ranked humans (\texttt{human\_048}, \texttt{human\_218}) and the
two top-ranked LLMs (\texttt{moonshotai/kimi-k2-0905},
\texttt{anthropic/claude-opus-4.7}) under the v5-2c AGC-Human
composite. Responses are unedited.

\paragraph{Alternate Uses Task --- prompt: BRICK.}
\label{tab:qual_aut}

\textbf{\texttt{human\_048}}: Rustic paperweight; Boho door stop; Visual
for class about updating your devices in a timely fashion; Visual when
explaining to one's parents why they cannot activate their HTC hero.

\textbf{\texttt{human\_218}}: Clean a pair of bricks and wrap them in
fabric, leather, or decorative paper to keep heavy hardcovers upright
on a shelf. Use a brick to hand-tamp soil or gravel for a small paver
project or a fence post hole. A single brick makes a heat-proof,
rustic base for thick pillar candles or incense burners.

\textbf{\texttt{moonshotai/kimi-k2-0905}}: Pocket-sized ``dumb-phone''
for digital detox vacations---just a brick you talk to and then feel
better about real life; official ``interruption veto'' gavel for Zoom
meetings---slam it on the desk to mute everyone instantly; DIY
home-planet kit: wrap it in aluminum foil, suspend it from the
ceiling, and watch your very own asteroid orbit the living room;
instant gravity-powered alarm clock---place it on a balanced shelf
above your bed; when the snooze ends the shelf tilts; reusable
dinner plate that guarantees no one asks for leftovers; pet rock's
aggressive cousin---set it in the garden to intimidate actual rocks
into growing flowers faster.

\textbf{\texttt{anthropic/claude-opus-4.7}}: Hollow it out and use it
as a rustic planter for succulents; freeze it and place it in a cooler
as a long-lasting ice block substitute; wrap it in festive paper as a
gag gift inside a larger package to confuse recipients; use it as a
pitching target in the backyard for baseball practice; grind it into
powder to make natural red pigment for paint or chalk; strap one to
each foot as makeshift resistance training for leg day.

Two stylistic registers are visible: human responses tend toward
idiomatic, casually-grounded phrasing (``HTC hero''), while LLM
responses tend toward literary or technical elaboration (``natural red
pigment for paint or chalk''). Both registers are legible as upper-tail
creative, but on different stylistic axes---consistent with the
cross-group production-strategy findings reported in
Appendix~\ref{app:agc_human_judge}. Verbatim responses for the
remaining four tasks (SCTT, Design Problems, Short Story, Metaphor
Completion) are released in the artifact bundle
(\texttt{agc\_human\_qualitative\_samples.csv}).

\end{document}